\documentclass{article}

\usepackage{arxiv}

\usepackage[utf8]{inputenc} 
\usepackage[T1]{fontenc}    
\usepackage[backref=page]{hyperref} 
\hypersetup{
    colorlinks=true,
    linkcolor=blue,
    filecolor=magenta,      
    urlcolor=cyan,
    citecolor=cyan,
    }
\usepackage{url}            
\usepackage{booktabs}       
\usepackage{amsfonts}       
\usepackage{nicefrac}       
\usepackage{microtype}      
\usepackage{amsmath} 
\usepackage{amssymb}
\usepackage{amsthm}
\usepackage{lipsum}
\usepackage{graphicx}
\usepackage{tikz} 
\usetikzlibrary{positioning, fit, backgrounds}
\usepackage{array}
\usepackage{subcaption}  
\usepackage{float}
\usepackage{cancel}
\usepackage{algorithm}
\usepackage{algpseudocode} 
\usepackage{array}
\usepackage{colortbl}
\usepackage{longtable}
\usepackage{enumitem}   
\usepackage{cite}

\theoremstyle{definition}

\usetikzlibrary{shapes.geometric, arrows.meta, positioning}

\graphicspath{ {./images/} }

\title{Aviary: training language agents on challenging scientific tasks}

\author{
\begin{tabular}{ccc}
Siddharth Narayanan$^{1}$ & James D. Braza$^{1}$ & Ryan-Rhys Griffiths$^{1}$ \\
Manu Ponnapati$^{1}$ & Albert Bou$^{1}$ & Jon Laurent$^{1}$\\
Ori Kabeli$^{1}$ & Geemi Wellawatte$^{1}$ & Sam Cox$^{1}$\\
Samuel G. Rodriques$^{1,3}$\thanks{These authors jointly supervise technical work at FutureHouse.} && Andrew D. White$^{1,2}$\footnotemark[1]
\end{tabular}
\\
\\
$^{1}$FutureHouse Inc., San Francisco, CA\\
$^{2}$University of Rochester, Rochester, NY\\
$^{3}$ Francis Crick Institute, London, UK\\
Correspondence to: \texttt{\{sam,andrew\}@futurehouse.org}
}

\begin{document}
\maketitle

\begin{abstract}
Solving complex real-world tasks requires cycles of actions and observations. 
This is particularly true in science, where tasks require many cycles of analysis, tool use, and experimentation. 
Language agents are promising for automating intellectual tasks in science because they can interact with tools via natural language or code. 
Yet their flexibility creates conceptual and practical challenges for software implementations, since agents may comprise non-standard components such as internal reasoning, planning, tool usage, as well as the inherent stochasticity of temperature-sampled language models. 
Here, we introduce Aviary, an extensible gymnasium for language agents. 
We formalize agents as policies solving language-grounded partially observable Markov decision processes, which we term language decision processes. 
We then implement five environments, including three challenging scientific environments: (1) manipulating DNA constructs for molecular cloning, (2) answering research questions by accessing scientific literature, and (3) engineering protein stability. 
These environments were selected for their focus on multi-step reasoning and their relevance to contemporary biology research. 
Finally, with online training and scaling inference-time compute, we show that language agents backed by open-source, non-frontier LLMs can match and exceed both frontier LLM agents and human experts on multiple tasks at up to 100x lower inference cost.
\end{abstract}

\section{Introduction}

\begin{figure}[ht]
    \centering
    \includegraphics[width=1\linewidth]{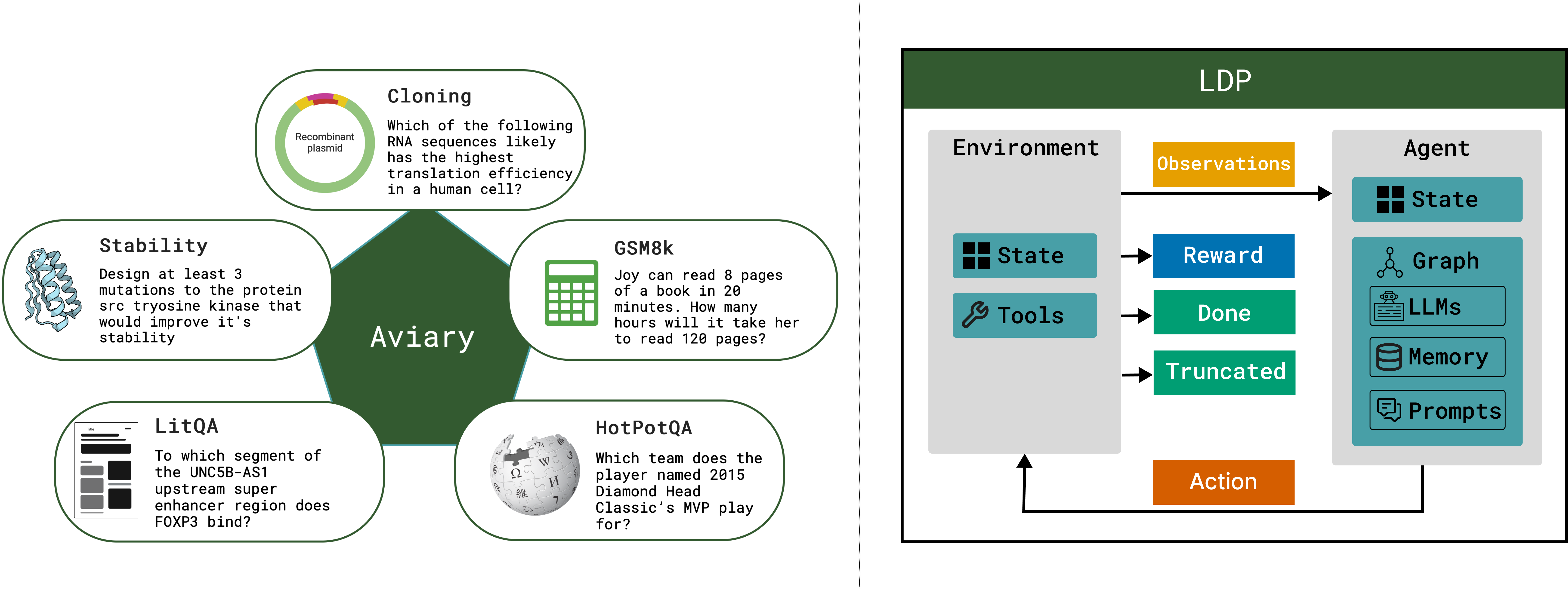}
    \caption{An overview of the five implemented Aviary environments and the language decision process (LDP) framework. The term language decision process here jointly refers to our theoretical description of the class of problems solved by language agents, as well as a software framework for implementing language agents based on a stochastic computation graph that enables training of language agent components such as LLM weights, prompts, memories, or LLM sampling parameters like temperature.}
    \label{fig:overview_schematic}
\end{figure}

Language agents~\cite{2023_Mialon, 2023_Xi, 2023_Gao, 2024_Sumers} are AI agents~\cite{2016_Russell} that integrate LLMs~\cite{2020_Brown, 2023_Achiam, 2023_Bowman} as core components. LLMs excel at zero-shot generalization~\cite{2022_Zeng, 2024_Szot}, providing a notable advantage over traditional AI agents, such as those based on handcrafted rules or reinforcement learning, which often struggle to generalize to new environments~\cite{2017_Lake}. While LLMs can exhibit flawed reasoning and logic when used in isolation~\cite{2022_Creswell, 2024_Frieder, 2024_Frieder2}, constructing a language agent by grounding LLMs in an environment with observational feedback can mitigate these issues. Early work on language agents used LLMs to directly output actions in the external environment~\cite{2023_Brohan, 2022_Huang, 2022_Dasgupta}, while more recently, language agents have been augmented with internal reasoning~\cite{2023_Yao, 2024_Shinn} and planning~\cite{2023_Hao, 2024_Yao} procedures, as well as long-term memory storage~\cite{2023_Park, 2024_Voyager}. 

An emergent research challenge is to pose a theoretical description of the learning problem solved by language agents~\cite{2024_Sumers, 2024_Zhuge} and to develop efficient methods to optimize the components of a language agent~\cite{2024_Zhuge, 2024_Yuksekgonul, 2024_Cheng_trace}. Here, we define common language agent tasks as language decision processes (LDPs) and frame language agents as stochastic computation graphs~\cite{2015_Schulman} that may be trained to solve LDPs. We show that pre-existing agents~\cite{2023_Yao, 2024_Shinn, 2024_Yao} can be implemented within our stochastic computation graph framework and introduce a simple and extensible software package named LDP that enables modular interchange of environments, agents, and optimizers, simplifying experimentation across a variety of settings.


In the problems we consider, we use the term optimization of language agents in the reinforcement sense to encompass procedures that yield iterative improvement of the language agent over time through feedback from an environment. An example of one such optimization algorithm is expert iteration (EI)~\cite{2017_Anthony, 2021_Anthony, 2024_Havrilla} which achieves learning through successive rounds of supervised fine-tuning on (self-) generated trajectories from a progressively stronger language agent. Other approaches include outcome supervision~\cite{uesato2022solving}, which is similar to rejection-sampled expert iteration, and process supervision~\cite{2024_Lightman}, which still selects high-reward trajectories, but also focuses on the individual steps of the outcome. 


%

In what follows, we introduce our definition of an environment, a language decision process, and optimization of agents backed by a stochastic computation graph. We recast popular benchmarks such as GSM8K~\cite{2021_Cobbe} and \textsc{hotpot}QA~\cite{2018_Yang} as environments and integrate three scientific environments related to challenging tasks in the natural sciences. The scientific environments are (1) DNA construct engineering, where the task is to answer questions pertaining to molecular cloning~\cite{laurent2024lab}, (2) scientific literature question answering, where the task is to answer a multiple choice question by finding a specific passage from the scientific literature~\cite{lala2023paperqa, skarlinski2024language}, and (3) protein design, where the goal is to propose mutations to improve the stability of a given protein sequence~\cite{Gao2020Deep,khakzad2023new}. On the DNA construct design and scientific literature question answering environments, we demonstrate that language agents based on the small, open-source \texttt{Llama-3.1-8B-Instruct} model, when trained with expert iteration and using inference-time majority vote sampling, can exceed the performance of both human experts and frontier LLMs.


The environment framework described in this work, Aviary, is available at \hyperlink{https://github.com/Future-House/aviary}{github.com/Future-House/aviary} and the stochastic computation graph framework together with language agent implementations and training code is available at \hyperlink{https://github.com/future-house/ldp}{github.com/Future-House/ldp}.
The components of and relationship between these frameworks is delineated in~\autoref{fig:overview_schematic}.

\section{Related Work}


\paragraph{Language Agent Formalisms}


Although language agents have achieved impressive empirical performance across a range of applications~\cite{2023_Mialon,skarlinski2024language,huang2024crispr}, there is still no universally agreed upon theoretical framework for defining a language agent. In terms of conceptual models, the cognitive architectures for language agents (CoALA) framework~\cite{2024_Sumers}, inspired by ideas from production systems and cognitive architectures, taxonomizes agents according to their information storage (working and long-term memories), decision-making procedures e.g. planning, and action space (divided into internal and external actions). Similarly, in \cite{2023_Weng_blog}, the author describes language agents as consisting of memory, planning, and tool usage components. Theoretically, many works represent language agents as partially observable Markov decision processes (POMDPs)~\cite{2023_Carta, 2023_Christianos, 2024_Wen, 2024_Wen2, 2024_Nguyen, 2024_Zhai, 2024_Song} yet differ in their treatment of the action space e.g. in \cite{2023_Christianos} the authors partition the action space into internal and external actions in a similar fashion to CoALA where internal actions are a family of functions that operate on the agent's memory and external actions elicit an interaction with the environment. By contrast, in \cite{2024_Wen} the authors do not make a distinction between internal and external actions. In~\cite{2024_Chen} the authors introduce a general framework for studying the design and analysis of LLM-based algorithms based on a computational graph where they assume LLM nodes are stateless, leaving consideration of aspects of language agents such as memory to future work.

\paragraph{Language Agent Optimization Frameworks}

Optimization of language agents may involve the learning of prompts, tool usage, LLM weights, LLM inference hyperparameters such as temperature, as well as more exotic language agent components such as edges between nodes in multiagent computation graphs. Frameworks such as LangChain~\cite{2022_Chase} and LlamaIndex~\cite{2022_Liu} support manual optimization of prompts via human editing. Optimizers such as EcoOptiGen~\cite{2023_Wang3} leverage black-box optimization schemes to learn LLM inference hyperparameters such as temperature, the maximum number of tokens, and the number of completions. Prompt optimization comprises the optimization of white-box LLMs and black-box LLMs (LLMs that exist behind an API and for which numerical gradients are unavailable). In white-box prompt optimization \cite{2020_Shin, 2021_Li, 2022_Jia, 2022_Chen3} numerical gradients can be taken over soft prompts \cite{2021_Qing}, the embedding representation of the text-based `hard' prompt. In black-box prompt optimization a multitude of techniques have been applied which attempt to overcome the absence of gradients \cite{2024_Guo, 2024_Ma2, 2024_Zhang, 2023_Cheng_Black, 2024_Yang, 2024_Lin_opt, 2024_Hu2, 2024_Wu2, 2024_Lin2, 2024_Chen2, 2023_Zhou, 2023_Pryzant, 2024_Sabbatella, 2023_Chen2, 2024_Wang3, 2024_Manas, 2023_Do, 2024_Sordoni, 2023_Sabbatella, 2024_Wen3, 2023_Ye, 2024_Wu3}. Tool learning \cite{2024_Qu, 2024_Schick} can be attempted through in-context demonstrations \cite{2024_Qin} or can seek to fine tune LLM weights on example demonstrations of appropriate tool usage~\cite{2024_Havrilla, 2024_Lumos} using techniques such as expert iteration~\cite{2017_Anthony, 2021_Anthony}. In terms of methods that seek to optimize many components of a language agent simultaneously, the TextGrad framework, introduced in \cite{2024_Yuksekgonul} backpropagates textual feedback received from an LLM. In a similar fashion, Zhou et. al \cite{2024_Zhou} also backpropagate textual feedback by creating natural language simulacrums of weights, losses, and gradients. In \cite{2024_Hu} the authors use a metaprompt to encourage an LLM to perform discrete optimization over an agent architecture. The Trace framework introduced in \cite{2024_Cheng_trace} proposes the OptoPrime optimizer which passes code execution traces in place of gradients and uses an LLM to provide textual feedback and perform updates. Another popular language agent optimization framework is DSPy \cite{2022_Khattab, 2023_Singhvi, 2024_Khattab} which parametrizes a computational graph for language agents and focuses on automatically generating and selecting useful demonstrations for in-context learning. In the multi-agent setting, GPTSwarm \cite{2024_Zhuge} introduces a computation graph and performs binary edge-level optimization and node-level optimization over prompts. Lastly, OpenR~\cite{2024_OpenR} is a framework for LLM reinforcement learning and inference-time scaling, but is targeted at token-level optimization, not tool usage.

\paragraph{Language Agent Benchmarks}

Existing language agent benchmarks feature a broad range of applications including machine learning tasks \cite{2024_Huang}, data science \cite{2024_Guo_ds, 2024_Grosnit}, data analysis \cite{2024_Hu3, 2024_Li}, quantitative reasoning \cite{2024_Liu}, and causal reasoning \cite{2023_Jin}. In Aviary, we place particular focus on scientific tasks. Relevant work in this area has included DiscoveryBench, a benchmark for data-driven hypothesis generation \cite{2024_Majumder}, ChemBench \cite{2024_Mirza} which focuses on chemistry tasks, BLADE \cite{2024_Blade} which is concerned with data-driven science, SciAgent \cite{2024_Ma} a benchmark for scientific reasoning, DISCOVERYWORLD \cite{2024_Jansen} which concentrates on cycles of scientific discovery, and ScienceWorld \cite{2022_Wang} which is concerned with scientific reasoning. For a review focused on scientifically-relevant agents the reader is directed to \cite{ramos2024review}. In Aviary, we focus on sequential decision-making tasks that necessitate multiple steps of agent-environment interactions. We construct environments from the pre-existing datasets such as GSM8K \cite{2021_Cobbe}, \textsc{hotpot}QA \cite{2018_Yang}, and LitQA2 \cite{skarlinski2024language} by casting them as parametrizable tools manipulating an environment state.


Our principal contributions are: (1) A precise definition of language decision processes (LDPs) for language-agent tasks and encompass many proposed agent architectures as stochastic computation graphs. (2) We introduce Aviary, a gym framework that emphasizes multi-step reasoning and tool usage, and provide five gym implementations (including three for scientific tasks). (3) We demonstrate that non-frontier LLMs, trained online with inference time sampling, can match or exceed the performance of frontier models on these tasks with a modest compute budget. (4) We release Aviary and our LDP framework as open-source software libraries to enable broader use and experimentation.

\section{Theory}

\subsection{Language Decision Processes}

A language decision process (LDP) is a Partially-Observable Markov Decision Process (POMDP)~\cite{1965_Aastrom} whose action and observation spaces are represented in natural language. More concretely, a LDP can be defined using the tuple $(\mathcal{V}, \mathcal{S}, \mathcal{A}, \mathcal{O}, T, Z, R, \gamma)$. 
Here, $\mathcal{V}$ is a non-empty alphabet\footnote{\label{vocab-footnote}In all LDPs we consider, $\mathcal{V}$ is the set of unicode characters, since Aviary is implemented in Python 3.}, $\mathcal{S}$ is the state space, $\mathcal{A} \subseteq \mathcal{V}^*$ is the action space\footnote{\label{kleene-footnote}Where $\mathcal{V^*} \stackrel{\text{def}}{=} \bigcup_{n=0}^{\infty} \mathcal{V}^n$ is the Kleene closure of a set $\mathcal{V}$~\cite{1956_Kleene, 2023_Meister}.}, $T(s'|s, a): \mathcal{S} \times \mathcal{A} \mapsto \mathcal{P}(\mathcal{S})$ is the transition function,  $R(s, a): \mathcal{S} \times \mathcal{A} \mapsto \mathcal{P}(\mathbb{R})$ is the reward function, $\mathcal{O} \subseteq \mathcal{V}^*$ is the observation space, $Z(o|s'): \mathcal{S} \times \mathcal{A} \mapsto \mathcal{P}(\mathcal{O})$ is the observation function\footnote{In all LDPs we consider, a state $s'\in \mathcal{S}$ uniquely defines an observation $o\in \mathcal{O}$. Unless otherwise specified, we omit the observation function $Z$.}, and $\gamma \in [0, 1]$ is the discount factor.


Unlike traditional reinforcement learning agents, feedback for language agents is ``grounded'' in the sense that environment observations must be converted to text~\cite{2024_Wu}. As such, the alphabet $\mathcal{V}$ is an important component of the LDP definition and follows other works~\cite{2023_Carta, 2024_Wen, 2024_Wen2}. The solution to an LDP is a policy $\pi_\theta: \mathcal{O} \mapsto \mathcal{A}$, where $\theta$ denotes the set of policy parameters we wish to learn. The parameter set $\theta$ is abstract and encapsulates any optimizable parameter of the language agent that may impact the action chosen such as LLM weights, inference hyperparameters such as temperature, as well as parametrized procedures such as internal reasoning. 


In contrast to previous works which demarcate between internal and external actions~\cite{2024_Sumers, 2023_Christianos}, where internal actions include reasoning and memory retrieval, in our problem framing we consider the action space to strictly constitute interactions with the external environment, allowing our parameter set $\theta$ to subsume optimizable procedures that are internal to the language agent such as memory retrieval and internal reasoning. Practically, it is worth noting that the complexity of our environments is such that we do not expect to obtain the globally optimal $\pi_\theta^*$. Our more modest goal is to be able to optimize $\theta$ in a direction that improves $\pi_\theta$ over time.

The observations in all environments we consider are deterministic functions of the state and so the reader may assume $Z = 1$ henceforth. For example, the environment may involve executing code and the observation is side-effects of the code. The state $\mathcal{S}$ would include all information necessary to induce the Markov property of the transition function: the file system, the versions of packages, any environment variables, the hardware, etc. However, the observation is just the output message of the executed code.

\subsection{Stochastic Computation Graphs}
\label{sec:scg}

In the general case, a language agent may include both stochastic and deterministic operations.
We build on the formalism of stochastic computation graphs (SCG)~\cite{2015_Schulman}: directed, acyclic graphs with nodes corresponding to computations and edges corresponding to arguments.

A deterministic node $v$ corresponds to a function $f_v$, and the node's output $o(v)$ is defined as:
\begin{equation}
    o(v) = f_v(\{o(w)\,|\,w\in\mathrm{parents}(v)\})
    \label{eq:det_output}    
\end{equation}
Similarly, a stochastic node $u$ is defined by a (conditional) distribution $p_u$, with output:
\begin{equation}
    o(u) \sim p_u(\,\cdot\,|\,\{o(w)\,|\,w\in\mathrm{parents}(v)\})
    \label{eq:stoch_output}
\end{equation}
Note that inputs to the graph can be treated as constant (deterministic) nodes.
Outputs of the graph are leaf nodes.

\begin{figure}[htb]
    \centering
    \begin{subfigure}[b]{0.3\textwidth}
        \centering
        \begin{tikzpicture}[scale=0.5]
            \tikzstyle{det} = [rectangle, draw, text centered, inner sep=3]
            \tikzstyle{stoch} = [rectangle, draw, dashed, text centered, inner sep=5]
            \tikzstyle{line} = [draw, -{Stealth[length=3mm, width=2mm]}]
        
            \node[det] (o) {$o_t$};
        
            \node[stoch, below=1cm of o] (lm) {$a_t\sim p_\mathrm{LLM}(\cdot \mid o_t)$};
            \path[line] (o) -- node[right] {} (lm);
        \end{tikzpicture}        
        \subcaption{Language model as policy}
        \label{fig:lm-policy}
    \end{subfigure}
    \hfill
    \begin{subfigure}[b]{0.3\textwidth}
        \centering
        \begin{tikzpicture}[scale=0.5]
            \tikzstyle{det} = [rectangle, draw, text centered, inner sep=3]
            \tikzstyle{stoch} = [rectangle, draw, dashed, text centered, inner sep=5]
            \tikzstyle{line} = [draw, -{Stealth[length=3mm, width=2mm]}]
        
            \node[det] (o) {$o_t$};
        
            \node[det, above right=0.7cm and 0.75cm of o] (knn) {kNN};
            \path[line] (o) -- node[right] {} (knn);
            
            \node[stoch, above right=0.1cm and 2.5cm of o] (lm) {$a_t\sim p_\mathrm{LLM}(\cdot \mid o_t, \text{KNN}(o_t))$};
            \path[line] (knn) -- node[right] {} (lm);
            \path[line] (o) -- node[right] {} (lm);
        \end{tikzpicture}
        \subcaption{Retrieval-augmented generation}
        \label{fig:rag}
    \end{subfigure}
    \hfill
    \vspace{1cm}

    \begin{subfigure}[b]{0.4\textwidth}
        \hspace{-0.5cm}
        \centering
        \begin{tikzpicture}[scale=0.5]
            \tikzstyle{det} = [rectangle, draw, text centered, inner sep=3]
            \tikzstyle{stoch} = [rectangle, draw, dashed, text centered, inner sep=5]
            \tikzstyle{line} = [draw, -{Stealth[length=3mm, width=2mm]}]
        
            \node[det] (o) {$o_t$};
    
            \node[stoch, above right=0.5 cm and 0.9 cm of o] (lm1) {$a_t^1\sim p_\mathrm{LLM}(\cdot \mid o_t)$};
            \path[line] (o) -- node[right] {} (lm1);
    
            \node[stoch, below right=0.5 cm and 0.9 cm of o] (lm2) {$a_t^2\sim p_\mathrm{LLM}(\cdot \mid o_t)$};
            \path[line] (o) -- node[right] {} (lm2);
    
            \node[det, right=4cm of o] (sel) {$a_t = \mathrm{argmax}\, q(o_t, a_t^i)$};
            \path[line] (lm1) -- node[right] {} (sel);
            \path[line] (lm2) -- node[right] {} (sel);
        \end{tikzpicture}
        \subcaption{Rejection sampling from language model}
        \label{fig:rejection-sampling}
    \end{subfigure}
    \hfill
    \begin{subfigure}[b]{0.4\textwidth}
        \centering
        \begin{tikzpicture}[scale=0.5]
            \tikzstyle{det} = [rectangle, draw, text centered, inner sep=3]
            \tikzstyle{stoch} = [rectangle, draw, dashed, text centered, inner sep=5]
            \tikzstyle{line} = [draw, -{Stealth[length=3mm, width=2mm]}]
        
            \node[det] (o) {$o_t$};
        
            \node[stoch, below=1cm of o] (lm1) {$x\sim p_\mathrm{LLM}(\cdot \mid o_t)$};
            \path[line] (o) -- node[right] {} (lm1);
            
            \node[stoch, below=1cm of lm1] (lm2) {$a_t\sim p_\mathrm{LLM}(\cdot \mid o_t, x)$};
            \path[line] (lm1) -- node[right] {} (lm2);
        \end{tikzpicture}        
        \subcaption{ReAct}
        \label{fig:react}
    \end{subfigure}
    
    \caption{Simple language agent architectures represented as stochastic computation graphs. Deterministic nodes are solid rectangles; stochastic nodes are dashed.
    Note that we augment the graphs with a deterministic input node to indicate how the observation $o_t$ is consumed.}
    \label{fig:la_as_scg}
\end{figure}
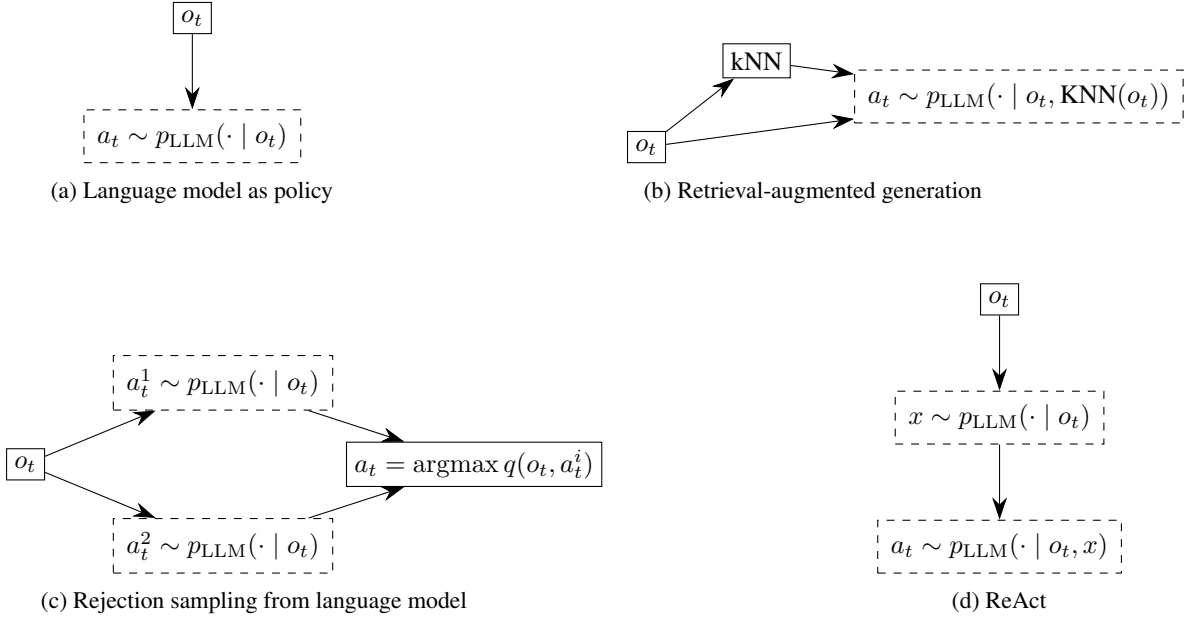

A language agent's policy is simply an SCG with a string input (the observation) and a string output (the action).
Language agent architectures can be easily expressed as SCGs by combining deterministic and stochastic nodes. The SCGs of the below common language agent architectures are visualized in~\autoref{fig:la_as_scg}. 

\begin{enumerate}
    \item[(a)] Language model as policy: a single stochastic node corresponding to sampling from the language model.
    \item[(b)] Retrieval-augmented generation (RAG): a deterministic node (document retrieval) leading to a stochastic node (LLM sampling).
    \item[(c)] Rejection sampling from LLM: several stochastic nodes (LLM samples), all leading to a deterministic node (selecting the preferred sample).
    \item[(d)] ReAct~\cite{2023_Yao}: two consecutive stochastic nodes, corresponding to sampling a reasoning string and an action (tool call).
\end{enumerate}

In many cases, a language agent defines an agent state $\xi_t$ that is a function of previous observations and actions. 
For example, the agent state of a multi-turn LLM conversation is typically defined as:
\begin{gather}
    \xi_t = [o_0, a_0, \dots, a_{t-1}, o_{t-1}] \nonumber \\ 
    a_t \sim p_\mathrm{LM}(\cdot | \xi_t)
\end{gather}
The SCG output is then a tuple of $(a_t, \xi_{t+1})$, namely an action and a new state. Separating the state enables batching of agent states and observations, as well as keeping the SCG as a true function. Memories, if desired, are considered part of the agent state, and their retrieval is incorporated in the SCG.

\subsection{Training Methods}

Below we describe commonly-used imitation learning \cite{1964_Widrow, 1969_Chambers, 1988_Pomerleau} methods employed to improve language agent performance on our environments. These training methods do not optimize the SCG graph directly, instead we optimize only the language model node in the SCG.


\paragraph{Behavior cloning (BC)} BC \cite{1990_Michie, 1995_Bain} refers to a general imitation learning technique that derives a policy by supervised learning on high quality trajectories termed expert demonstrations. In the context of language agents this is typically achieved by supervised fine-tuning (SFT) of an LLM on either human trajectories or trajectories generated from a stronger LLM \cite{2023_Christianos, 2023_Shunyu, 2023_Zeng, 2024_Lumos, 2024_Song}. In the context of our experiments, we use BC to initialize the trajectory buffer for an expert iteration loop on \texttt{Llama-3.1-8B-Instruct}, due to its inability to self-generate successful trajectories prior to training.

\label{sec:ei}

\paragraph{Expert iteration (EI)} EI~\cite{2017_Anthony, 2021_Anthony, 2024_Havrilla} performs behavior cloning in an iterative fashion, improving the demonstration data each iteration. The inputs to the EI algorithm are a base LLM represented as an initial policy $\pi_0$ and a trajectory buffer $D_0$, which may either be empty or consist of an initial set of demonstrations generated by a human expert or a stronger LLM.
At each round of EI, first a batch $B$ of trajectories are sampled from the current policy $\pi_i$ (via rollout). Then these trajectories $\{\tau_i^{(j)}\}_{j=1}^{B}$ are filtered (the rejection sampling step~\cite{2023_Yuan2}) based on return $R$ exceeding a threshold value $\rho$. The filtered trajectories are then appended to the trajectory buffer $D_i$ and the current LLM, $\pi_i$, is fine-tuned on $D_i$ using cross-entropy loss. Pseudocode for EI is provided in Algorithm~1.


\begin{algorithm}
\caption{Expert Iteration with Rejection Sampling Fine-Tuning}
\label{alg:expert_iteration}
\begin{algorithmic}[1]
\State \textbf{Inputs}: initial policy $\pi_0$, iteration rounds $N$, batch size $B$, return threshold $\rho$, trajectory buffer $D_0$
\For{$i = 1, \dots N$}
    \State $T_i \gets \text{rollout}(\pi_{i-1})$ 
    \State $D_i \gets D_{i-1} \cup \{ (\tau_i^{(j)}, R_i^{(j)}) |\ \tau_i^{(j)} \in T_i, R_i^{(j)} > \rho, j = 1, \dots, B \}$ \Comment{Rejection sample trajectories}
    \State $\pi_i \gets \text{SFT}(D_i)$ \Comment{SFT on updated trajectory buffer}
\EndFor
\end{algorithmic}
\end{algorithm}

\paragraph{Inference Compute Scaling} Scaling inference-time compute to improve LLM performance is now a frequently-employed technique \cite{brown2024large, wang2022self}. There are two common settings: oracle-verified (pass@$k$) and majority vote (consensus@$k$). As shown in Brown et al.\cite{brown2024large}, if an oracle verifier can identify any correct solution -- namely, if you can obtain just one correct answer among $k$ -- then it is possible to scale across multiple orders of magnitude. Without an oracle verifier, majority voting can be used\cite{wang2022self}. Majority voting is simply the consensus response, which requires some natural binning of responses. Although oracle verification scales to very large numbers of completions\cite{li2022competition}, majority voting plateaus more quickly than oracle verification\cite{brown2024large}. In this work, we omit any unsure or truncated trajectories (trajectories for which the agent did not submit an answer) from majority voting.

\section{Environments}
\label{sec:envs}

We briefly detail the environments comprising Aviary. Further details on the environments may be found in the appendix.

\subsection{GSM8K}

The GSM8K environment is based on the GSM8K dataset introduced in \cite{2021_Cobbe}, which consists of linguistically diverse grade school math word problems designed to assess multi-step mathematical reasoning. The GSM8K dataset comprises a training set of 7,473 questions and a test set of 1,319 questions.
The environment exposes a calculator tool.





\subsection{\textsc{hotpot}QA}

The \textsc{hotpot}QA environment is based on the \textsc{hotpot}QA dataset introduced in \cite{2018_Yang}, which was subsequently extended to a language agent environment in \cite{2023_Yao}. The \textsc{hotpot}QA dataset comprises 112,779 question-answer pairs. We run evals on the 7,405 eval subset of questions. In the \textsc{hotpot}QA environment, the agent is provided with a Wikipedia API and tasked with answering the questions. There are is no given context to the agent and the API supports access to all of Wikipedia articles and sections.

\subsection{PaperQA}

PaperQA~\cite{lala2023paperqa, skarlinski2024language} is a language agent/environment pairing developed for literature research and question answering that leverages reranking and contextual summarization. Specifically in \cite{skarlinski2024language}, an untrained version 2 of PaperQA, called PaperQA2, attained superhuman-level precision and human-level accuracy on version 2 of a literature question and answer task, called LitQA2~\cite{laurent2024lab}. PaperQA2 was implemented with tools and a tool calling agent, so we refactored PaperQA2 to be an Aviary environment as part of the version 5 release of the \href{https://pypi.org/project/paper-qa/}{\texttt{paper-qa} Python package}. To make it easy for the machine learning community to use, we modified the search tool to center on local storage containing a set of PDF, text, and HTML files using tantivy~\cite{tanitvy-code}. This local search is why we call this PaperQA variant ``PaperQA2 Local.'' The citation traversal tool was omitted for this local setting. A \texttt{complete} tool was added to support agents that require at least one tool selection and allow the agent to declare if the answer addresses all parts of the question. 

LitQA2 features 248 questions, 199 of which are publicly available and the remaining 49 were held out as a test set. We reuse the same test set here for comparability with~\cite{skarlinski2024language}. The remaining 199 questions were randomly 80\%-20\% split such that the training set is 149 questions and the evaluation set is 40 questions. The test split questions can be found in the \href{https://huggingface.co/datasets/futurehouse/aviary-paper-data}{\texttt{aviary-paper-data} Hugging Face dataset}. Note this PaperQA environment is capable of doing tasks beyond LitQA2. For example, it can do literature review writing and contradiction detection as reported in Skarlinski et. al~\cite{skarlinski2024language}.

To build the search indexes, we (1) aggregated all paper search or citation traversal results from a database of run logs made during~\cite{skarlinski2024language}, (2) binned the results corresponding to each LitQA2 question, and (3) combined bins based our train, evaluation, and test split LitQA2 questions. The end result is 18955 DOIs in the train split, 5457 DOIs in the evaluation split, and 5519 DOIs in the test split. The train, evaluation, and test split DOIs are can be found in the \href{https://github.com/Future-House/paper-qa}{\texttt{paper-qa} GitHub repository}. Each split has over 100X the reachable DOIs compared to contained questions so agents face a learnable retrieval task. Due to copyright of the underlying papers, we only distribute their DOIs, not the parsed text used by the environment index.

\subsection{Molecular Cloning}

Molecular cloning is a fundamental technique of manipulating DNA in biomedical research, enabling a majority of basic research such gene function studies, creating transgenic models, and producing recombinant proteins~\cite{Bertero2017Methods,Sharma2014Advances}. The molecular cloning process results in a DNA ``construct,'' which is a general term for DNA that encodes for the desired biologic molecule or genes. Molecular cloning involves assembling DNA fragments, ligating them into vectors, introducing the recombinant DNA into host organisms, and screening for desired clones~\cite{Bertero2017Methods}. The steps in molecular cloning are usually done with a combination of human planning, specialized software, and databases of known purchasable components.

We have formulated this into an environment. The molecular cloning environment is composed of the main tools used by experts in the lab: (1) an annotation tool that can predict the function of segments of a plasmid (2) a natural language search tool that retrieves sequences given text and (3) tools required to plan the protocols. The protocol specific tools include PCR primer design, ligation, codon optimization, Gibson or Golden gate assembly, and fetching genes from standard organisms. Many implementations use or were derived from the \href{https://github.com/bebop/poly}{Go poly library}. The annotation tools were built using MMSeqs2~\cite{steinegger2017mmseqs2}. The complete list of tools is given in the supporting information.

The specific tasks used for evaluation come from the SeqQA benchmark~\cite{laurent2024lab}. SeqQA consists of ``textbook'' style questions, such as counting the fragments after digestion, predicting translated sequences, and identifying polymerase chain reaction primers. A complete description of their construction and human evaluation is given in Laurent et. al \cite{laurent2024lab}. The test SeqQA questions in this work have not been previously released, but are created using the same procedure and are available in the \href{https://huggingface.co/datasets/futurehouse/aviary-paper-data}{\texttt{aviary-paper-data} Hugging Face dataset}. There are 150 test questions, although we omit 10 related to RNA specifically. There are 500 train questions, which we take from the original SeqQA release accessible \href{https://huggingface.co/datasets/futurehouse/lab-bench/viewer/SeqQA}{here}. SeqQA is solvable with only a subset of the tools, and the tools have \textit{not} been engineered specifically for the conventions of SeqQA. For example, SeqQA questions assume 1-indexing, but the tools are 0-indexed and thus the language agent needs to learn to convert. Another example is that SeqQA only considers coding open reading frames, but the tools can consider both reading frames.

The combination of tools for manipulating DNA constructs, annotating sequences or plasmids, and searching for DNA components in databases enables tasks beyond SeqQA. The environment also supports working with ``CloningScenarios,'' which is a more multiple-choice benchmark for working with DNA constructs derived from real lab notebooks\cite{laurent2024lab}. One can also do normal plasmid tasks, such as ``Clone the given protein [protein] into [plasmid] to express in yeast with a GFP fusion (check annotations above, plus GFP in correct relative orientation).''


\subsection{Protein Stability}

Engineering proteins with increased stability is an essential task in protein engineering, with wide-ranging applications in enzyme engineering and drug design \cite{sheldon2018role}. Protein stability is a general term for a protein's ability to retain function under non-native conditions, such as increased temperature, lowered pH, or aggregation-inducing solvents. Numerous sequence-based and structure-based approaches have been developed to enhance protein stability \cite{Goldenzweig2018-rm}, including deep learning methods such as ThermoMPNN \cite{thermompnn}. However, protein stability is determined by complex protein sequence and structure properties along with biological context making it challenging to predict accurately with existing in-silico approaches \cite{BROOM2020717}. Therefore, an approach that integrates protein structure and sequence methods, including physics-based methods like Rosetta, can provide a more comprehensive understanding of biophysical determinants of protein stability \cite{MAESTRO}. 

The protein stability environment is composed of tools commonly used by human experts to analyze a protein sequence and structure. The main tools are (1)a biochemical description tool,that describes the types of bonds between any residues in the protein sequence,; 2) a sequence property description tool that describes the molecular weight, aromaticity, instability index, iso-electric point, sequence charge, and average hydropathy of a protein sequence; 3) a secondary structure annotation tool; and (4) a Rosetta tool to compute aggregation propensity score per residue \cite{sapscore}.

 310.ai \cite{310ai} introduced a chat interface for natural language-based protein design, though it is not an agent. Conversely, methods such as ProteinForceGPT, an autonomous large language model (LLM) agent introduced by \cite{ghafarollahi_protagents_2024}, uses pre-trained models to predict force–separation curves, supplemented by models like Chroma~\cite{ingraham2023illuminating} and OmegaFold~\cite{Wu2022.07.21.500999}. Furthermore, recent work has shown the effectiveness of LLM as biological sequence optimizers \cite{chen2024llmshighlyconstrainedbiophysicalsequence}. Our environment offers a framework for training of agents that can effectively integrate knowledge from physics-based models, biochemical principles, and pre-trained protein models while leveraging experimental results to improve protein stability.

We assess the language agent's performance on 40 proteins randomly selected from the megascale protein stability dataset, excluding any that are mentioned in the text of \cite{Tsuboyama2023}. Proposed mutations are evaluated using the Rosetta cart\_ddg protocol~\cite{Frenz2020}.


\section{Results}

We assess the capabilities of tool-equipped language agents to solve problems in the aforementioned environments. These environments require iterative cycles of tool calls and observation. 
We then explore behavior cloning and expert iteration to train agents on specific tasks in environments. Finally, we explore using inference-time compute via majority sampling to improve performance.

\begin{figure}[ht]
    \centering
    \includegraphics[width=1\linewidth]{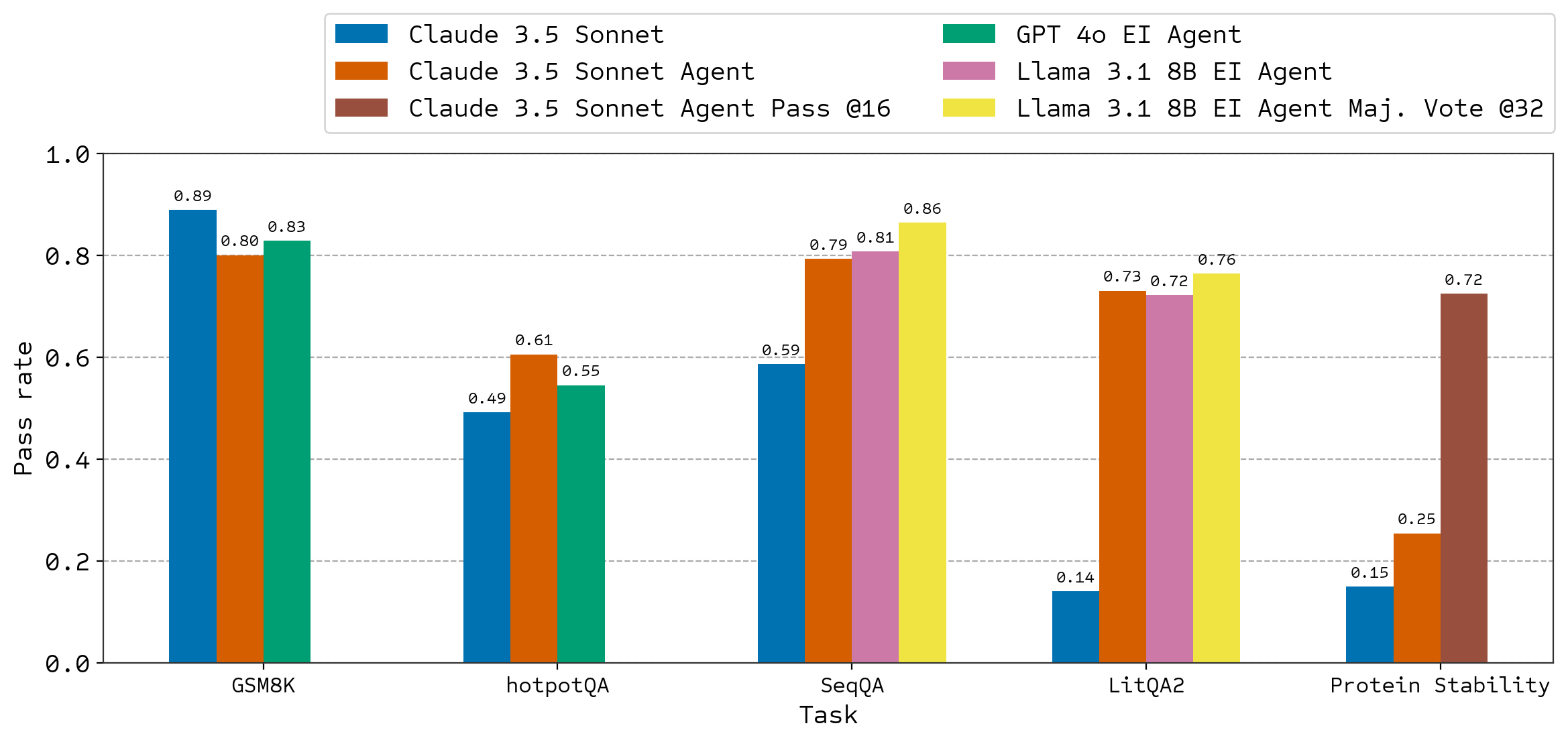}
    \caption{
    Ability of LLMs and language agents to solve tasks using Aviary environments.
    All LDP-trained agents are optimized using behavior cloning and expert iteration.
    For GSM8K and \textsc{hotpot}QA, EI is performed on \texttt{gpt-4o};
    SeqQA and LitQA2 use \texttt{Llama-3.1-8B-Instruct} (see~\autoref{fig:ei-results}).
    The difference in GSM8K zero-shot reported here (89\%) vs Anthropic benchmarks~\cite{anthropic2024claude3} (96.5\%) is likely because Anthropic's use of chain-of-thought prompting, which we did not use. All agents are rolled out on the environment for a maximum of 10 steps, with the exception of PaperQA, which allowed up to 18 steps.
    }
    \label{fig:baseline_pass_rate}
\end{figure}

An overview of the models used in this work and their performance on our tasks is shown in~\autoref{fig:baseline_pass_rate}. This includes both trained (described below) and frontier language models. They are:
\begin{itemize}
    \item Zero-shot Claude 3.5 Sonnet: \texttt{claude-3-5-sonnet-20241022} is prompted to solve the tasks without access to the environment or any tools. No example output or formatting instructions are given. 
    \item Claude 3.5 Sonnet agent: A language agent that prompts \texttt{claude-3-5-sonnet-20241022} to call environment tools until the task is solved. It uses the recommended Anthropic API tool-calling schema\cite{anthropic2024claude3}. The observation emitted from a tool call may include a summary or details about the environment state.
    \item LDP-trained language agent: An LDP agent that has been trained to solve tasks using the environment. This can either be based on fine-tuned \texttt{gpt-4o} (GSM8K, hotpotQA) or \texttt{Llama-3.1-8B-Instruct} (SeqQA, LitQA2).
    \item Majority voting: We sample 32 trajectories using the trained LDP agent and use their consensus as the solution to the task. For protein stability, we do oracle-verification/pass@k, as in protein engineering one typically tests a batch and only keep the most successful. \cite{brown2024large}
\end{itemize}

The mixture of existing benchmarks and closed-source models was chosen to demonstrate the flexibility of the Aviary software. Claude 3.5 Sonnet was the best frontier LLM we tested across tasks, and was thus used as the benchmark for comparison. With the exception of GSM8K, all agents are able to improve over the zero-shot baseline when given access to the environment. In the case of GSM8K, we hypothesize that a sequence of calculator calls (with no intermediate reasoning) is out-of-distribution with respect to the LLMs' training data, which may also contain math word problems.
This is consistent with recent findings~\cite{mirzadeh2024}, where modifying elements of the original questions or adding irrelevant information also caused performance degradation across multiple models, as such changes similarly introduce a distribution shift from the training data.

Training LDP agents improves performance over untrained LDP agents of the same architecture. On challenging tasks (SeqQA, LitQA2), a relatively small model (\texttt{Llama-3.1-8B-Instruct}) can be trained to match performance of a much larger frontier model (\texttt{claude-3-5-sonnet}). Majority voting can be used to sample multiple times from the LDP agents, giving a further large gain at the cost of increased inference compute. The protein stability task sees a large improvement for pass@16, which is a well-known effect for oracle-verified problems\cite{brown2024large}. These results are broken out in more detail below.

\subsection{Behavior Cloning and Expert Iteration}
\label{fig:ei-results}
Using \texttt{ldp}, we train language agents in the environments described above.
Since these environments are challenging, expert iteration initially rejects the majority of trajectories, leading to very slow learning.
We therefore begin with a period of behavior cloning, using high-quality trajectories collected by rejection-sampling from a larger LLM.
Once the language agent can solve a reasonable fraction of training problems, we switch to expert iteration.
All experiments are conducted with \texttt{Llama-3.1-8B-Instruct}~\cite{grattafiori2024llama3herdmodels} as the base language model, using Nvidia A100 GPUs. 

\begin{figure}[h!]
    \centering
    \includegraphics[width=1\linewidth]{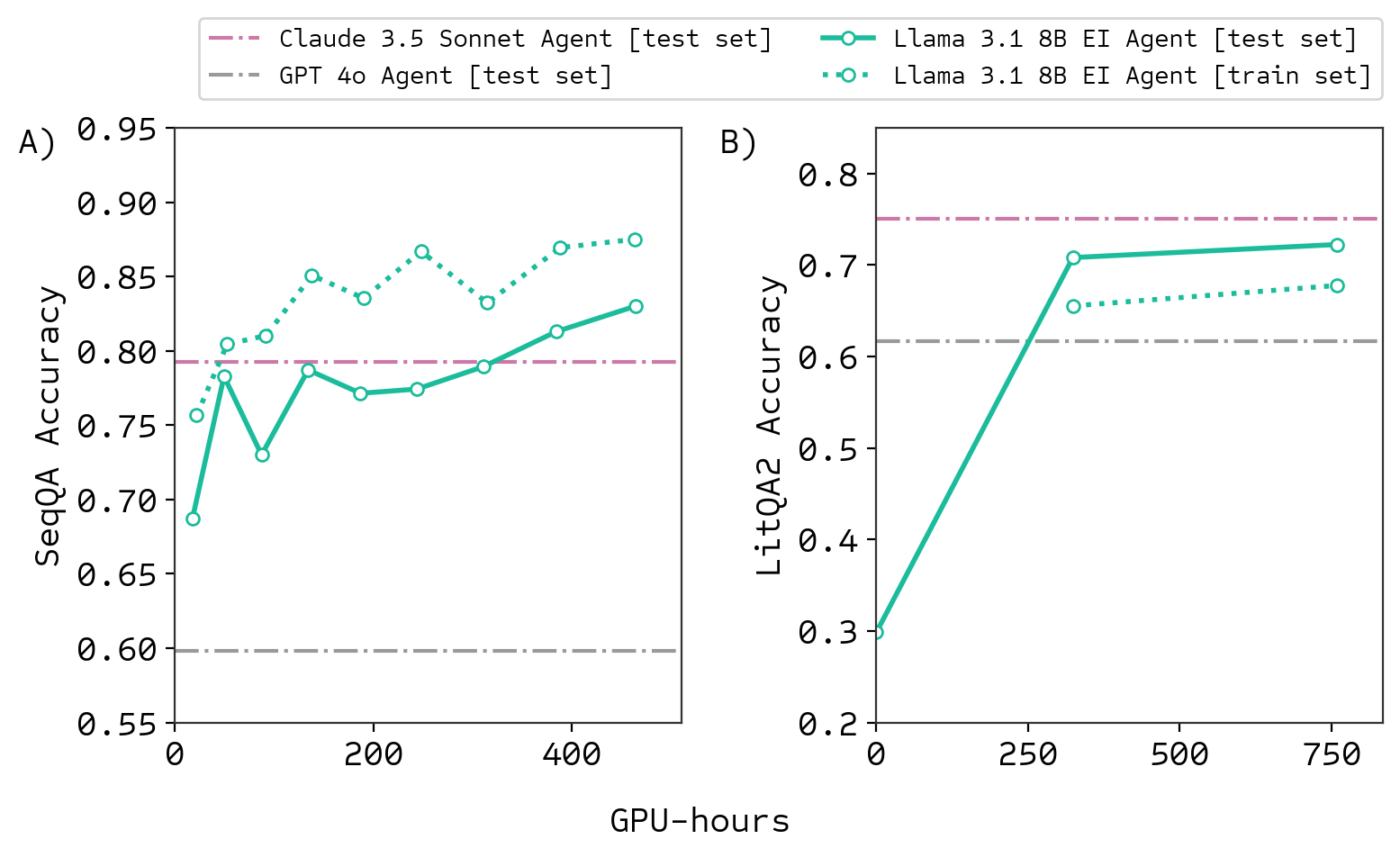}
    \caption{
        Training language agents to solve (A) SeqQA tasks using the molecular cloning environment and (B) LitQA2 problems using the PaperQA environment.
        The first epoch represents behavior cloning, followed by expert iteration.
        These experiments use multiple workers to asynchronously collect trajectories and train the model, so \texttt{GPU-hours} measures the total time spent sampling and training.
        An untrained \texttt{Llama-3.1-8B-Instruct} agent solves 1\% of SeqQA tasks, so we omit the data point at \texttt{GPU-hours=0} in panel A.
    }
    \label{fig:ei-acc-curves}
\end{figure}

In~\autoref{fig:ei-acc-curves}A, we show the results of training an agent (\texttt{Llama-3.1-8B} EI) to solve SeqQA tasks using the molecular cloning environment.
Expert iteration is seeded with 2841 valid trajectories (behavior cloning), followed by 8 further EI epochs using trajectories from the agent as it improves.
Behavior cloning provides a large initial jump in performance, with a further 14\% (absolute) improvement from online learning.
We note a gap in performance on train and test set tasks, indicating some degree of overfitting. 

In~\autoref{fig:ei-acc-curves}B, we show the results of a similar procedure applied to LitQA2 problems in the PaperQA environment.
In this case, the untrained \texttt{Llama-3.1-8B} agent has non-trivial performance (30\% accuracy), but still significantly improves from behavior cloning (430 trajectories).
Because some LitQA2 problems are easier than others, we want to focus training on difficult tasks.
Therefore, during expert iteration, we sample trajectories from each task in the dataset with probability:
\begin{equation}
    P(\text{task }k) = \frac{w_k}{\sum_j w_j}; \quad w_k = M\cdot(1-f_\mathrm{pass}^k)
\end{equation}
where $f_\mathrm{pass}^k$ is a moving average of task $k$'s pass rate as the agent is trained and $M$ is a scaling factor (set to 20).
With this, EI produces a small improvement beyond behavior cloning, up to 72\% on the test set. 

\begin{figure}[h!]
    \centering
    \includegraphics[width=1\linewidth]{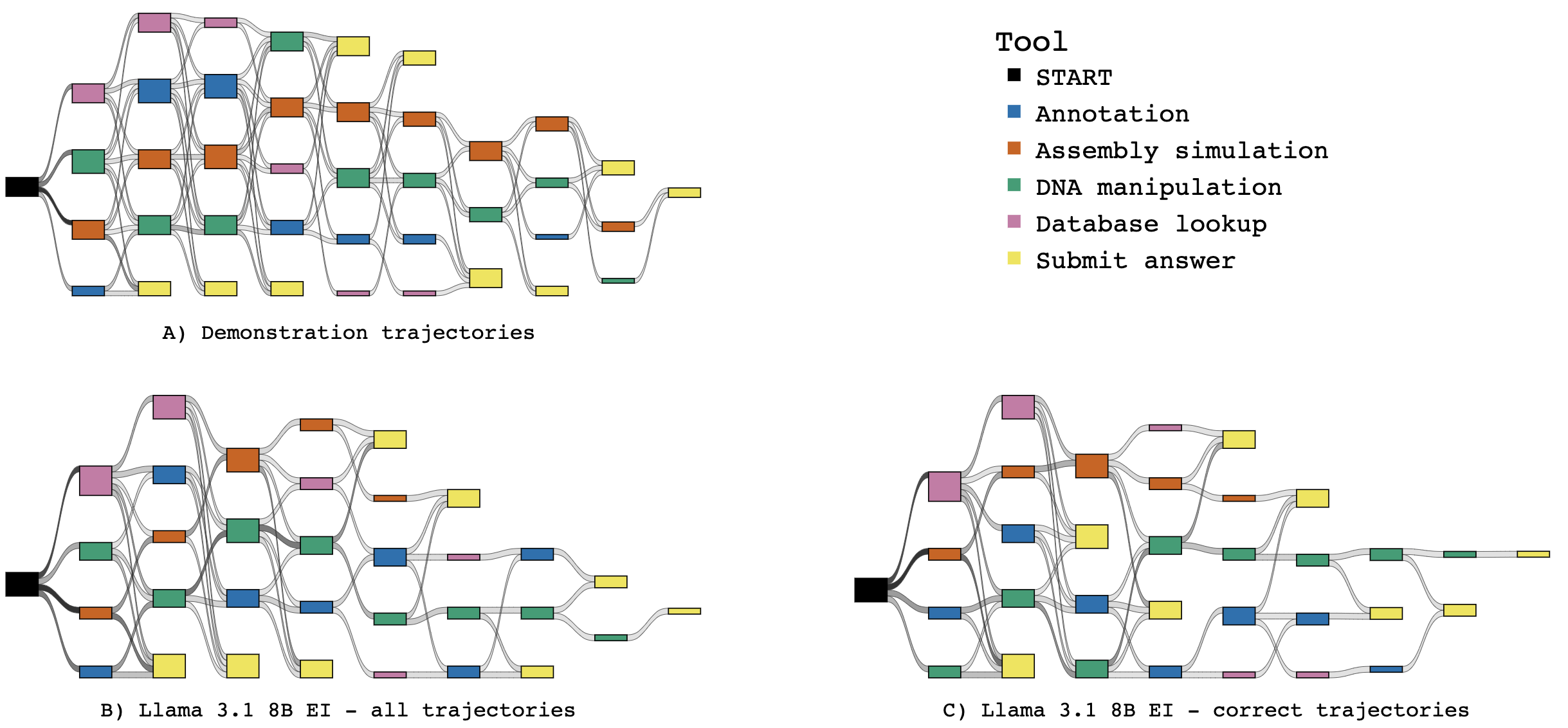}
    \caption{
        Patterns of tool calls across trajectories.
        Colored boxes represent different tool categories, and edges between boxes represent consecutive actions taken in a trajectory, with darker edges implying more trajectories.
        In panel (A), we show the demonstration trajectories used for behavior cloning.
        Panels (B) and (C) show the trajectories sampled from the \texttt{Llama-3.1-8B} EI agent after expert iteration. 
    }
    \label{fig:seqqa-sankey}
\end{figure}

Finally, in ~\autoref{fig:seqqa-sankey}, we study the distribution of SeqQA trajectories explored by a trained language agent. The demonstration trajectories (all correct) heavily feature assembly simulations and are relatively long. 
The trained agent was initially cloned from the demonstrations, but through online learning discovered significantly different ways to solve SeqQA tasks. 
Its trajectories are generally shorter and less diverse, suggesting that self-training tends to converge on a subset of possible paths.

\subsection{Inference Compute Scaling}

\begin{figure}[h!]
    \centering
    \includegraphics[width=\linewidth]{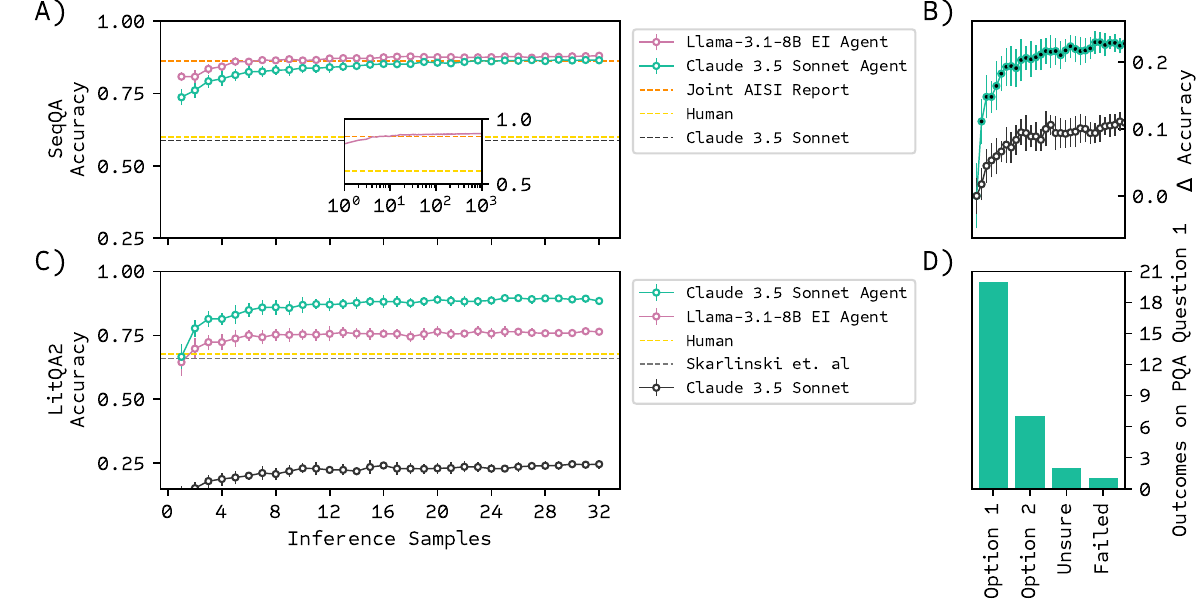}
    \caption{A) majority voting accuracy in SeqQA setting as a function of sampled trajectories for trained \texttt{Llama-3.1-8B} EI agent and \texttt{claude-3-5-sonnet-241022} agent. The majority voting increases performance to exceed previously reported scaffolded agents on this task (Joint AISI Report)\cite{joint_aisi}, which used \texttt{claude-3-5-sonnet-241022}. ``Claude 3.5 Sonnet'' is \texttt{claude-3-5-sonnet-241022} without tools (simply answering questions directly). The inset shows further gains from 0.86 to 0.89 from 32 to 945 samples. B) shows the change in accuracy of an LLM without tools vs. the language agent on SeqQA. The language agent has a greater gain from majority voting, showing >0.2 gain. C) majority voting accuracy on LitQA2 in PaperQA2 environment. Both agents significantly exceed previously measured human performance and best previous models on this task\cite{skarlinski2024language}. Claude 3.5-Sonnet Agent plateaus at 90\% accuracy. D) Shows an example question voting on LitQA2 (question id \texttt{3e6d7a54}). Option 1 is the correct response, option 2 is incorrect, and failure is because the agent did not submit an answer prior to trajectory termination. Error bars in the plots are computed by bootstrap resampling.}
    \label{fig:test} 
\end{figure}

We assessed majority voting on two of the environments that have multiple choice answers -- SeqQA and LitQA2 -- to see if it improves benchmarks in the LDP setting. We evaluated on test splits that we neither trained on nor should be in frontier LLM training corpus because they are not on the public internet. Figure~\ref{fig:test} shows the results on these two environments. Majority voting generally gives large improvements for these agents - about 20 percentage points of accuracy on both environments. Figure~\ref{fig:test}B shows that the LLM, without tools, gets about a 10 percentage point gain (although from a much lower starting accuracy). Figure~\ref{fig:test}D shows an example of what the majority votes look like on one specific LitQA2 question. Option 1 is correct, option 2 is a minority option and sometimes chosen, there is an unsure option, and lastly any truncated trajectories (e.g. unexpected failure during rollout or hitting max allowable steps). For majority voting, we filter out the unsure and failed trajectories. The biggest gains in accuracy are from 0 to 4 samples, but it continues up to 32 samples. These two environments are less than 200 unique questions, so the trend may continue if we have a larger set of unique questions. 

Figure~\ref{fig:test}C shows that majority voting on LitQA2 with a Claude 3.5 Sonnet agent gives 0.89 accuracy on the test set, significantly exceeding previously reported scores of 0.67 from Skarlinski et. al \cite{skarlinski2024language} and human performance reported in Laurent et. al\cite{laurent2024lab}. The \texttt{Llama-3.1-8B} EI agent shows good performance, matching human and previously reported best at only one sample. Three samples exceeds those marks, but it cannot match the Sonnet agent if it also uses majority voting on more than one sample. Nevertheless, exceeding a frontier LLM in the single sample setting on unseen data with a small model is a surprising result. 

Figure~\ref{fig:test}A shows that consensus sampling of the \texttt{Llama-3.1-8B} EI agent significantly exceeds a Claude 3.5 Sonnet agent at all sample counts. SeqQA is more structured than LitQA2, requiring more consistent and longer tool call sequences. The \texttt{Llama-3.1-8B} EI agent can be sampled from cheaply, and so we ran 945 rollouts (Figure~\ref{fig:test}A inset). We still observe gains out to 100s of samples, giving a final accuracy of 0.89. The highest previously reported result on SeqQA was from a joint technical report from the US and UK AI Safety Institutes on pre-deployment evaluation of \texttt{claude-3.5-sonnet-20241022} at 0.87 accuracy.

\subsection{Inference Cost Scaling}

The results of the previous sections demonstrate how the performance of different agents scales as training time and sampled trajectories are increased.
In this section, we offer a more practical metric: inference cost.
This becomes especially relevant in a high-throughput setting, in which agents are tasked to solve thousands of problems in parallel. 

\begin{figure}[h!]
    \centering
    \includegraphics[width=\linewidth]{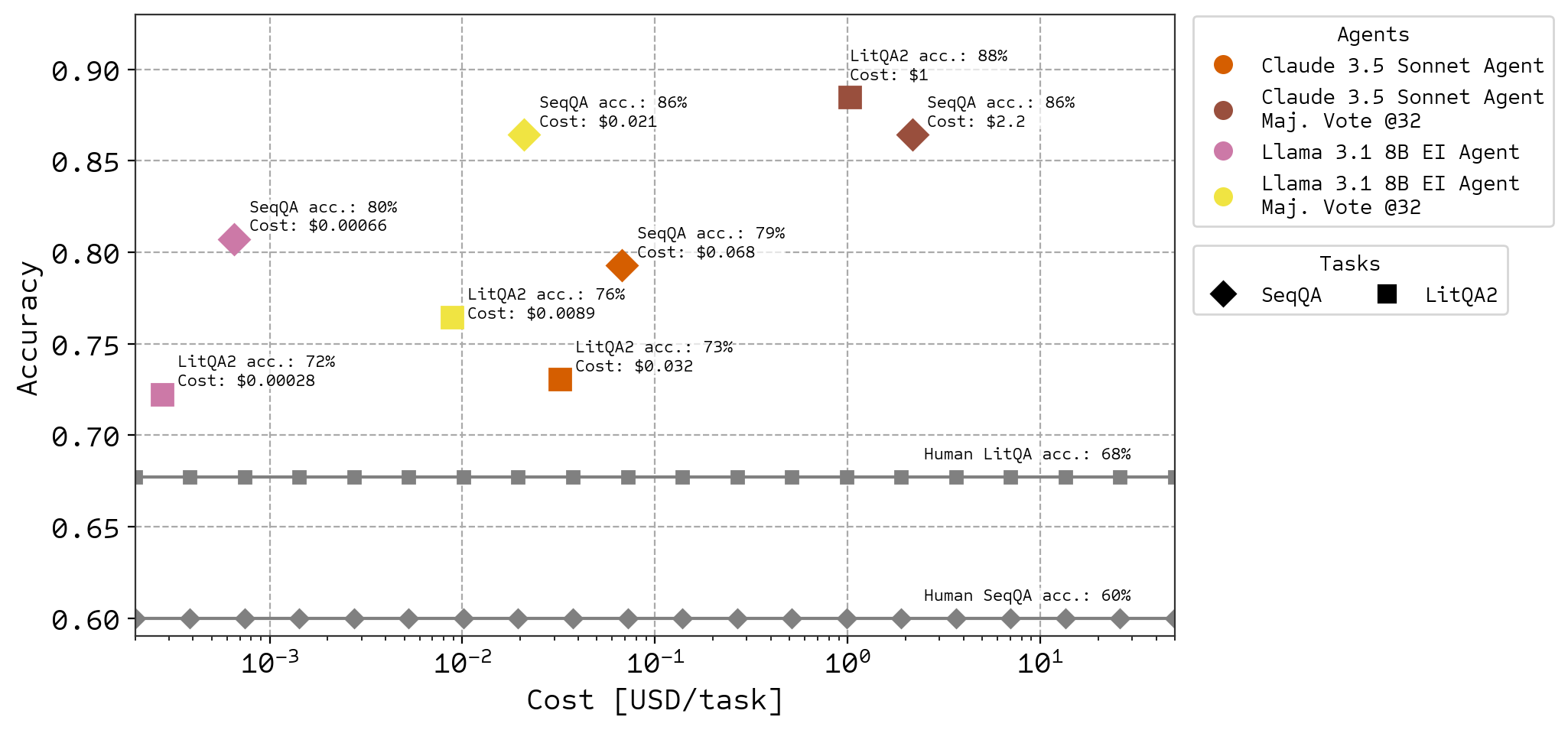}
    \caption{Accuracy vs. inference cost of two agents (Claude 3.5 Sonnet and \texttt{Llama-3.1-8B} EI) on two sets of tasks (SeqQA and LitQA2).
    For both agents, both the single-rollout and majority-voting settings are considered.
    Note that all inference settings here outperform human performance (reported in~\cite{skarlinski2024language} and~\cite{laurent2024lab}).}
    \label{fig:cost-scaling}
\end{figure}

We focus our comparison on the Claude 3.5 Sonnet agent versus the \texttt{Llama-3.1-8B} EI agent.
We use the following rates for LLM inference at time of writing:
\begin{itemize}
    \item Claude 3.5 Sonnet: \$3/1M input tokens and \$15/1M output tokens\footnote{\href{https://www.anthropic.com/pricing\#anthropic-api}{https://www.anthropic.com/pricing\#anthropic-api}}.
    \item \texttt{Llama-3.1-8B}: \$0.03/1M input and output tokens, a price typical in the LLM inference market\footnote{\href{https://lambdalabs.com/inference\#pricing}{https://lambdalabs.com/inference\#pricing}}. While this price refers to inference of the vanilla model, we use it as a reasonable estimate for serving a EI-trained model with a similar set of optimizations.
\end{itemize}
In~\autoref{fig:cost-scaling}, we report performance and inference cost on SeqQA and LitQA2.
While majority voting with the Claude 3.5 Sonnet agent clearly outperforms other settings, this requires $\mathcal{O}(\$1)$ per task.
We reach the same SeqQA accuracy using the \texttt{Llama-3.1-8B} EI agent for 100x less cost.
While this was not achievable for LitQA2, we note that majority voting with \texttt{Llama-3.1-8B} EI still exceeds single-rollout with Sonnet for 3x less cost.

\section{Discussion}

We have presented a framework of a language agent and environment interacting to solve tasks that require multiple steps of reasoning and tool usage, which we call a language decision process. We implemented five environments, including three related to biology problems. These three environments contain a variety of tasks, but we focus on tasks with benchmarks that are easy to evaluate, namely LitQA and SeqQA (multiple choice) as well as a task focussed on modifying enzymes to improve their stability. Language agents in these environments perform significantly better than non-agentic LLMs.

We have applied two methods to improve the performance of language agents on the tasks we consider: expert iteration and majority voting. Smaller models, such as \texttt{Llama-3.1-8B-Instruct}, perform poorly without additional training. The application of behavior cloning and expert iteration can overcome this limitation, however, achieving comparable performance to the best evaluated agents based on black-box frontier models (in this case \texttt{claude-3-5-sonnet-241022}). Majority voting provides additional performance gains at inference time. Majority voting benefits all agents and non-agentic LLMs tested, though it provides more benefit to agents. This is likely due to the fact that environment trajectories possess elevated level of stochasticity relative to a completion sampled from a non-agentic LLM. The protein stability task also benefits from increasing inference compute, though it uses oracle-verification. 

Using an 8B model reduces the inference costs such that it is feasible to vote amongst $\sim$1000 rollouts per task. For reference, our SeqQA tasks require 7-10 LLM calls (Figure~\ref{fig:seqqa-sankey}) and cost \$0.07 on average per trajectory with \texttt{claude-3-5-sonnet-241022} and \$0.00066 per trajectory at current pricing with \texttt{Llama-3.1-8B} EI (although the tasks could be run for free on many consumer laptops). The human PhD contractors that represent the human data series in Figure~\ref{fig:test} cost between \$4 and \$12 per question (see Laurent et. al \cite{laurent2024lab} for a description of the combined hourly, performance, and completion bonuses). In summary, trained agents can exceed the accuracy of human and frontier models at 100x cheaper cost.

There are some limitations in this work. The comparisons are not exactly matched between humans and other previous reported results on benchmarks. One reason is that the Environments are complex pieces of software, with dozens of dependencies that all have specific versions. Another issue is that the splits we used are only now available (although they were included in the human assessments). We used splits that were impossible to have been scraped in training data. One reason we took this precaution is that websites have re-hosted the LitQA2 benchmark without the canary string\cite{srivastava2023beyond}, so that it is plausibly now in LLM pre-training corpuses.

Comparing against humans is also fraught, because they do not have access to the same tools. Although, in Laurent et. al \cite{laurent2024lab} there were incentives for correct answers, ample time, and only restrictions against using AI tools. Nevertheless, it's always possible that the humans could have been given better or more precise technology for the task. Ultimately, the test of these language agents is their ability to make novel scientific discoveries and not getting high scores on benchmarks.

\section{Conclusion}

We have presented Aviary, a gymnasium for language agents. Aviary currently contains five environments, three of which focus on challenging scientific tasks. Language agents, implemented in these environments, exceed the performance of zero-shot frontier LLMs on the SeqQA, \textsc{hotpot}QA, LitQA2, and protein stability tasks. Language agents also exceed human performance on SeqQA and LitQA2.

We have introduced the language decision process (LDP) framework for formally describing language agent tasks and showed that language agents can be cast as stochastic computation graphs. Through behavior cloning, expert iteration, and inference-time sampling, we demonstrated that trained \texttt{Llama-3.1-8B} EI agents can match and exceed the performance of humans and frontier LLMs in the LitQA2 and SeqQA benchmarks at significantly lower cost. Thus, we have demonstrated that modest compute budgets and model sizes can be competitive at solving realistic scientific tasks. The reported trained \texttt{Llama-3.1-8B} EI agents are compute efficient and exceed human-level task performance, enabling high-throughput automation of meaningful scientific tasks across biology.

Both the Aviary (\href{https://github.com/Future-House/aviary}{\texttt{aviary}}) and LDP (\href{https://github.com/Future-House/ldp}{\texttt{ldp}}) frameworks are open source and should serve as useful libraries for implementing environments and language agents.

\section*{Acknowledgments}
Work at FutureHouse is supported by the generosity of Eric and Wendy Schmidt. The results and models reported in this work used compute resources from the National AI Research Resource Pilot, including support from NVIDIA and the NVIDIA DGX Cloud. We also acknowledge all members of FutureHouse for useful discussions, including Cade Gordon, Peter Chang, Michael Skarlinski, and Conor Igoe.

\bibliographystyle{unsrt}  
\bibliography{references}

\begin{thebibliography}{100}

\bibitem{2023_Mialon}
Gr{\'e}goire Mialon, Roberto Dessi, Maria Lomeli, Christoforos Nalmpantis, Ramakanth Pasunuru, Roberta Raileanu, Baptiste Roziere, Timo Schick, Jane Dwivedi-Yu, Asli Celikyilmaz, et~al.
\newblock Augmented language models: a survey.
\newblock {\em Transactions on Machine Learning Research}, 2023.

\bibitem{2023_Xi}
Zhiheng Xi, Wenxiang Chen, Xin Guo, Wei He, Yiwen Ding, Boyang Hong, Ming Zhang, Junzhe Wang, Senjie Jin, Enyu Zhou, et~al.
\newblock The rise and potential of large language model based agents: A survey.
\newblock {\em arXiv preprint arXiv:2309.07864}, 2023.

\bibitem{2023_Gao}
Chen Gao, Xiaochong Lan, Nian Li, Yuan Yuan, Jingtao Ding, Zhilun Zhou, Fengli Xu, and Yong Li.
\newblock Large language models empowered agent-based modeling and simulation: A survey and perspectives.
\newblock {\em arXiv preprint arXiv:2312.11970}, 2023.

\bibitem{2024_Sumers}
Theodore Sumers, Shunyu Yao, Karthik Narasimhan, and Thomas Griffiths.
\newblock Cognitive architectures for language agents.
\newblock {\em Transactions on Machine Learning Research}, 2024.
\newblock Survey Certification.

\bibitem{2016_Russell}
Stuart~J Russell and Peter Norvig.
\newblock {\em Artificial intelligence: a modern approach}.
\newblock Pearson, 2016.

\bibitem{2020_Brown}
Tom Brown, Benjamin Mann, Nick Ryder, Melanie Subbiah, Jared~D Kaplan, Prafulla Dhariwal, Arvind Neelakantan, Pranav Shyam, Girish Sastry, Amanda Askell, et~al.
\newblock Language models are few-shot learners.
\newblock {\em Advances in Neural Information Processing Systems}, 33:1877--1901, 2020.

\bibitem{2023_Achiam}
Josh Achiam, Steven Adler, Sandhini Agarwal, Lama Ahmad, Ilge Akkaya, Florencia~Leoni Aleman, Diogo Almeida, Janko Altenschmidt, Sam Altman, Shyamal Anadkat, et~al.
\newblock {GPT-4 Technical Report}.
\newblock {\em arXiv preprint arXiv:2303.08774}, 2023.

\bibitem{2023_Bowman}
Samuel~R Bowman.
\newblock Eight things to know about large language models.
\newblock {\em arXiv preprint arXiv:2304.00612}, 2023.

\bibitem{2022_Zeng}
Andy Zeng, Maria Attarian, Krzysztof~Marcin Choromanski, Adrian Wong, Stefan Welker, Federico Tombari, Aveek Purohit, Michael~S Ryoo, Vikas Sindhwani, Johnny Lee, et~al.
\newblock Socratic models: Composing zero-shot multimodal reasoning with language.
\newblock In {\em The Eleventh International Conference on Learning Representations}, 2023.

\bibitem{2024_Szot}
Andrew Szot, Max Schwarzer, Harsh Agrawal, Bogdan Mazoure, Rin Metcalf, Walter Talbott, Natalie Mackraz, R~Devon Hjelm, and Alexander~T Toshev.
\newblock Large language models as generalizable policies for embodied tasks.
\newblock In {\em The Twelfth International Conference on Learning Representations}, 2024.

\bibitem{2017_Lake}
Brenden~M Lake, Tomer~D Ullman, Joshua~B Tenenbaum, and Samuel~J Gershman.
\newblock Building machines that learn and think like people.
\newblock {\em Behavioral and Brain Sciences}, 40:e253, 2017.

\bibitem{2022_Creswell}
Antonia Creswell, Murray Shanahan, and Irina Higgins.
\newblock Selection-inference: exploiting large language models for interpretable logical reasoning.
\newblock In {\em The Eleventh International Conference on Learning Representations}, 2023.

\bibitem{2024_Frieder}
Simon Frieder, Luca Pinchetti, Ryan-Rhys Griffiths, Tommaso Salvatori, Thomas Lukasiewicz, Philipp Petersen, and Julius Berner.
\newblock Mathematical capabilities of {ChatGPT}.
\newblock {\em Advances in Neural Information Processing Systems}, 36, 2024.

\bibitem{2024_Frieder2}
Simon Frieder, Jonas Bayer, Katherine~M Collins, Julius Berner, Jacob Loader, Andr{\'a}s Juh{\'a}sz, Fabian Ruehle, Sean Welleck, Gabriel Poesia, Ryan-Rhys Griffiths, et~al.
\newblock Data for mathematical copilots: Better ways of presenting proofs for machine learning.
\newblock {\em arXiv preprint arXiv:2412.15184}, 2024.

\bibitem{2023_Brohan}
Anthony Brohan, Yevgen Chebotar, Chelsea Finn, Karol Hausman, Alexander Herzog, Daniel Ho, Julian Ibarz, Alex Irpan, Eric Jang, Ryan Julian, et~al.
\newblock {Do as I can, not as I say:} grounding language in robotic affordances.
\newblock In {\em Conference on Robot Learning}, pages 287--318. PMLR, 2023.

\bibitem{2022_Huang}
Wenlong Huang, Pieter Abbeel, Deepak Pathak, and Igor Mordatch.
\newblock Language models as zero-shot planners: Extracting actionable knowledge for embodied agents.
\newblock In {\em International Conference on Machine Learning}, pages 9118--9147. PMLR, 2022.

\bibitem{2022_Dasgupta}
Ishita Dasgupta, Christine Kaeser-Chen, Kenneth Marino, Arun Ahuja, Sheila Babayan, Felix Hill, and Rob Fergus.
\newblock Collaborating with language models for embodied reasoning.
\newblock In {\em NeurIPS 2022 Foundation Models for Decision Making Workshop}, 2022.

\bibitem{2023_Yao}
Shunyu Yao, Jeffrey Zhao, Dian Yu, Nan Du, Izhak Shafran, Karthik Narasimhan, and Yuan Cao.
\newblock {ReAct}: synergizing reasoning and acting in language models.
\newblock In {\em International Conference on Learning Representations (ICLR)}, 2023.

\bibitem{2024_Shinn}
Noah Shinn, Federico Cassano, Ashwin Gopinath, Karthik Narasimhan, and Shunyu Yao.
\newblock Reflexion: Language agents with verbal reinforcement learning.
\newblock {\em Advances in Neural Information Processing Systems}, 36, 2024.

\bibitem{2023_Hao}
Shibo Hao, Yi~Gu, Haodi Ma, Joshua Hong, Zhen Wang, Daisy Wang, and Zhiting Hu.
\newblock Reasoning with language model is planning with world model.
\newblock In {\em Proceedings of the 2023 Conference on Empirical Methods in Natural Language Processing}, pages 8154--8173, 2023.

\bibitem{2024_Yao}
Shunyu Yao, Dian Yu, Jeffrey Zhao, Izhak Shafran, Tom Griffiths, Yuan Cao, and Karthik Narasimhan.
\newblock Tree of thoughts: Deliberate problem solving with large language models.
\newblock {\em Advances in Neural Information Processing Systems}, 36, 2024.

\bibitem{2023_Park}
Joon~Sung Park, Joseph O'Brien, Carrie~Jun Cai, Meredith~Ringel Morris, Percy Liang, and Michael~S Bernstein.
\newblock Generative agents: Interactive simulacra of human behavior.
\newblock In {\em Proceedings of the 36th Annual ACM Symposium on User Interface Software and Technology}, pages 1--22, 2023.

\bibitem{2024_Voyager}
Guanzhi Wang, Yuqi Xie, Yunfan Jiang, Ajay Mandlekar, Chaowei Xiao, Yuke Zhu, Linxi Fan, and Anima Anandkumar.
\newblock Voyager: An open-ended embodied agent with large language models.
\newblock {\em Transactions on Machine Learning Research}, 2024.

\bibitem{2024_Zhuge}
Mingchen Zhuge, Wenyi Wang, Louis Kirsch, Francesco Faccio, Dmitrii Khizbullin, and J{\"u}rgen Schmidhuber.
\newblock {GPTSwarm}: Language agents as optimizable graphs.
\newblock In {\em Forty-first International Conference on Machine Learning}, 2024.

\bibitem{2024_Yuksekgonul}
Mert Yuksekgonul, Federico Bianchi, Joseph Boen, Sheng Liu, Zhi Huang, Carlos Guestrin, and James Zou.
\newblock {TextGrad: Automatic" differentiation" via text}.
\newblock {\em arXiv preprint arXiv:2406.07496}, 2024.

\bibitem{2024_Cheng_trace}
Ching-An Cheng, Allen Nie, and Adith Swaminathan.
\newblock {Trace is the New AutoDiff--unlocking efficient optimization of computational workflows}.
\newblock {\em arXiv preprint arXiv:2406.16218}, 2024.

\bibitem{2015_Schulman}
John Schulman, Nicolas Heess, Theophane Weber, and Pieter Abbeel.
\newblock Gradient estimation using stochastic computation graphs.
\newblock {\em Advances in Neural Information Processing Systems}, 28, 2015.

\bibitem{2017_Anthony}
Thomas Anthony, Zheng Tian, and David Barber.
\newblock Thinking fast and slow with deep learning and tree search.
\newblock In {\em Proceedings of the 31st International Conference on Neural Information Processing Systems}, pages 5366--5376, 2017.

\bibitem{2021_Anthony}
Thomas~William Anthony.
\newblock {\em Expert iteration}.
\newblock PhD thesis, UCL (University College London), 2021.

\bibitem{2024_Havrilla}
Alex Havrilla, Yuqing Du, Sharath~Chandra Raparthy, Christoforos Nalmpantis, Jane Dwivedi-Yu, Maksym Zhuravinskyi, Eric Hambro, Sainbayar Sukhbaatar, and Roberta Raileanu.
\newblock Teaching large language models to reason with reinforcement learning.
\newblock {\em arXiv preprint arXiv:2403.04642}, 2024.

\bibitem{uesato2022solving}
Jonathan Uesato, Nate Kushman, Ramana Kumar, Francis Song, Noah Siegel, Lisa Wang, Antonia Creswell, Geoffrey Irving, and Irina Higgins.
\newblock Solving math word problems with process-and outcome-based feedback.
\newblock {\em arXiv preprint arXiv:2211.14275}, 2022.

\bibitem{2024_Lightman}
Hunter Lightman, Vineet Kosaraju, Yuri Burda, Harrison Edwards, Bowen Baker, Teddy Lee, Jan Leike, John Schulman, Ilya Sutskever, and Karl Cobbe.
\newblock Let's verify step by step.
\newblock In {\em The Twelfth International Conference on Learning Representations}, 2024.

\bibitem{2021_Cobbe}
Karl Cobbe, Vineet Kosaraju, Mohammad Bavarian, Mark Chen, Heewoo Jun, Lukasz Kaiser, Matthias Plappert, Jerry Tworek, Jacob Hilton, Reiichiro Nakano, et~al.
\newblock Training verifiers to solve math word problems.
\newblock {\em arXiv preprint arXiv:2110.14168}, 2021.

\bibitem{2018_Yang}
Zhilin Yang, Peng Qi, Saizheng Zhang, Yoshua Bengio, William Cohen, Ruslan Salakhutdinov, and Christopher~D Manning.
\newblock {HotpotQA}: A dataset for diverse, explainable multi-hop question answering.
\newblock In {\em Proceedings of the 2018 Conference on Empirical Methods in Natural Language Processing}, pages 2369--2380, 2018.

\bibitem{laurent2024lab}
Jon~M Laurent, Joseph~D Janizek, Michael Ruzo, Michaela~M Hinks, Michael~J Hammerling, Siddharth Narayanan, Manvitha Ponnapati, Andrew~D White, and Samuel~G Rodriques.
\newblock {LAB-Bench}: Measuring capabilities of language models for biology research.
\newblock {\em arXiv preprint arXiv:2407.10362}, 2024.

\bibitem{lala2023paperqa}
Jakub L{\'a}la, Odhran O'Donoghue, Aleksandar Shtedritski, Sam Cox, Samuel~G Rodriques, and Andrew~D White.
\newblock {PaperQA}: Retrieval-augmented generative agent for scientific research.
\newblock {\em arXiv preprint arXiv:2312.07559}, 2023.

\bibitem{skarlinski2024language}
Michael~D Skarlinski, Sam Cox, Jon~M Laurent, James~D Braza, Michaela Hinks, Michael~J Hammerling, Manvitha Ponnapati, Samuel~G Rodriques, and Andrew~D White.
\newblock Language agents achieve superhuman synthesis of scientific knowledge.
\newblock {\em arXiv preprint arXiv:2409.13740}, 2024.

\bibitem{Gao2020Deep}
Wenhao Gao, Sai~Pooja Mahajan, Jeremias Sulam, and Jeffrey~J. Gray.
\newblock Deep learning in protein structural modeling and design.
\newblock {\em Patterns}, 1(9):100142, December 2020.

\bibitem{khakzad2023new}
Hamed Khakzad, Ilia Igashov, Arne Schneuing, Casper Goverde, Michael Bronstein, and Bruno Correia.
\newblock A new age in protein design empowered by deep learning.
\newblock {\em Cell Systems}, 14(11):925--939, 2023.

\bibitem{huang2024crispr}
Kaixuan Huang, Yuanhao Qu, Henry Cousins, William~A Johnson, Di~Yin, Mihir Shah, Denny Zhou, Russ Altman, Mengdi Wang, and Le~Cong.
\newblock {CRISPR-GPT}: An {LLM} agent for automated design of gene-editing experiments.
\newblock {\em arXiv preprint arXiv:2404.18021}, 2024.

\bibitem{2023_Weng_blog}
Lilian Weng.
\newblock {LLM}-powered autonomous agents.
\newblock {\em lilianweng.github.io}, Jun 2023.

\bibitem{2023_Carta}
Thomas Carta, Cl{\'e}ment Romac, Thomas Wolf, Sylvain Lamprier, Olivier Sigaud, and Pierre-Yves Oudeyer.
\newblock Grounding large language models in interactive environments with online reinforcement learning.
\newblock In {\em International Conference on Machine Learning}, pages 3676--3713. PMLR, 2023.

\bibitem{2023_Christianos}
Filippos Christianos, Georgios Papoudakis, Matthieu Zimmer, Thomas Coste, Zhihao Wu, Jingxuan Chen, Khyati Khandelwal, James Doran, Xidong Feng, Jiacheng Liu, et~al.
\newblock {Pangu-Agent: A fine-tunable generalist agent with structured reasoning}.
\newblock {\em arXiv preprint arXiv:2312.14878}, 2023.

\bibitem{2024_Wen}
Muning Wen, Ziyu Wan, Weinan Zhang, Jun Wang, and Ying Wen.
\newblock Reinforcing language agents via policy optimization with action decomposition.
\newblock {\em arXiv preprint arXiv:2405.15821}, 2024.

\bibitem{2024_Wen2}
Muning Wen, Cheng Deng, Jun Wang, Weinan Zhang, and Ying Wen.
\newblock Entropy-regularized token-level policy optimization for large language models.
\newblock {\em arXiv preprint arXiv:2402.06700}, 2024.

\bibitem{2024_Nguyen}
Dang Nguyen, Viet~Dac Lai, Seunghyun Yoon, Ryan~A Rossi, Handong Zhao, Ruiyi Zhang, Puneet Mathur, Nedim Lipka, Yu~Wang, Trung Bui, et~al.
\newblock {DynaSaur}: Large language agents beyond predefined actions.
\newblock {\em arXiv preprint arXiv:2411.01747}, 2024.

\bibitem{2024_Zhai}
Yuanzhao Zhai, Tingkai Yang, Kele Xu, Feng Dawei, Cheng Yang, Bo~Ding, and Huaimin Wang.
\newblock Enhancing decision-making for {LLM} agents via step-level q-value models.
\newblock {\em arXiv preprint arXiv:2409.09345}, 2024.

\bibitem{2024_Song}
Yifan Song, Da~Yin, Xiang Yue, Jie Huang, Sujian Li, and Bill~Yuchen Lin.
\newblock Trial and error: Exploration-based trajectory optimization for {LLM} agents.
\newblock {\em arXiv preprint arXiv:2403.02502}, 2024.

\bibitem{2024_Chen}
Yanxi Chen, Yaliang Li, Bolin Ding, and Jingren Zhou.
\newblock {On the Design and Analysis of LLM-Based Algorithms}.
\newblock {\em arXiv preprint arXiv:2407.14788}, 2024.

\bibitem{2022_Chase}
Harrison Chase.
\newblock {LangChain}, October 2022.

\bibitem{2022_Liu}
Jerry Liu.
\newblock {LlamaIndex}, November 2022.

\bibitem{2023_Wang3}
Chi Wang, Xueqing Liu, and Ahmed~Hassan Awadallah.
\newblock Cost-effective hyperparameter optimization for large language model generation inference.
\newblock In {\em International Conference on Automated Machine Learning}, pages 21--1. PMLR, 2023.

\bibitem{2020_Shin}
Taylor Shin, Yasaman Razeghi, Robert~L Logan~IV, Eric Wallace, and Sameer Singh.
\newblock {AutoPrompt}: Eliciting knowledge from language models with automatically generated prompts.
\newblock In {\em Proceedings of the 2020 Conference on Empirical Methods in Natural Language Processing (EMNLP)}, pages 4222--4235, 2020.

\bibitem{2021_Li}
Xiang~Lisa Li and Percy Liang.
\newblock Prefix-tuning: Optimizing continuous prompts for generation.
\newblock In {\em Proceedings of the 59th Annual Meeting of the Association for Computational Linguistics and the 11th International Joint Conference on Natural Language Processing (Volume 1: Long Papers)}, pages 4582--4597, 2021.

\bibitem{2022_Jia}
Menglin Jia, Luming Tang, Bor-Chun Chen, Claire Cardie, Serge Belongie, Bharath Hariharan, and Ser-Nam Lim.
\newblock Visual prompt tuning.
\newblock In {\em European Conference on Computer Vision}, pages 709--727. Springer, 2022.

\bibitem{2022_Chen3}
Xiang Chen, Ningyu Zhang, Xin Xie, Shumin Deng, Yunzhi Yao, Chuanqi Tan, Fei Huang, Luo Si, and Huajun Chen.
\newblock Knowprompt: Knowledge-aware prompt-tuning with synergistic optimization for relation extraction.
\newblock In {\em Proceedings of the ACM Web conference 2022}, pages 2778--2788, 2022.

\bibitem{2021_Qing}
Guanghui Qin and Jason Eisner.
\newblock Learning how to ask: Querying lms with mixtures of soft prompts.
\newblock In {\em Proceedings of the 2021 Conference of the North American Chapter of the Association for Computational Linguistics: Human Language Technologies}, pages 5203--5212, 2021.

\bibitem{2024_Guo}
Qingyan Guo, Rui Wang, Junliang Guo, Bei Li, Kaitao Song, Xu~Tan, Guoqing Liu, Jiang Bian, and Yujiu Yang.
\newblock Connecting large language models with evolutionary algorithms yields powerful prompt optimizers.
\newblock In {\em The Twelfth International Conference on Learning Representations}, 2024.

\bibitem{2024_Ma2}
Ruotian Ma, Xiaolei Wang, Xin Zhou, Jian Li, Nan Du, Tao Gui, Qi~Zhang, and Xuanjing Huang.
\newblock Are large language models good prompt optimizers?
\newblock {\em arXiv preprint arXiv:2402.02101}, 2024.

\bibitem{2024_Zhang}
Tuo Zhang, Jinyue Yuan, and Salman Avestimehr.
\newblock {Revisiting OPRO: The Limitations of Small-Scale LLMs as Optimizers}.
\newblock {\em arXiv preprint arXiv:2405.10276}, 2024.

\bibitem{2023_Cheng_Black}
Jiale Cheng, Xiao Liu, Kehan Zheng, Pei Ke, Hongning Wang, Yuxiao Dong, Jie Tang, and Minlie Huang.
\newblock Black-box prompt optimization: Aligning large language models without model training.
\newblock {\em arXiv preprint arXiv:2311.04155}, 2023.

\bibitem{2024_Yang}
Chengrun Yang, Xuezhi Wang, Yifeng Lu, Hanxiao Liu, Quoc~V Le, Denny Zhou, and Xinyun Chen.
\newblock Large language models as optimizers.
\newblock In {\em The Twelfth International Conference on Learning Representations}, 2024.

\bibitem{2024_Lin_opt}
Xiaoqiang Lin, Zhongxiang Dai, Arun Verma, See-Kiong Ng, Patrick Jaillet, and Bryan Kian~Hsiang Low.
\newblock Prompt optimization with human feedback.
\newblock {\em arXiv preprint arXiv:2405.17346}, 2024.

\bibitem{2024_Hu2}
Wenyang Hu, Yao Shu, Zongmin Yu, Zhaoxuan Wu, Xiangqiang Lin, Zhongxiang Dai, See-Kiong Ng, and Bryan Kian~Hsiang Low.
\newblock Localized zeroth-order prompt optimization.
\newblock {\em arXiv preprint arXiv:2403.02993}, 2024.

\bibitem{2024_Wu2}
Zhaoxuan Wu, Xiaoqiang Lin, Zhongxiang Dai, Wenyang Hu, Yao Shu, See-Kiong Ng, Patrick Jaillet, and Bryan Kian~Hsiang Low.
\newblock Prompt optimization with {EASE?} {Efficient} ordering-aware automated selection of exemplars.
\newblock {\em arXiv preprint arXiv:2405.16122}, 2024.

\bibitem{2024_Lin2}
Xiaoqiang Lin, Zhaoxuan Wu, Zhongxiang Dai, Wenyang Hu, Yao Shu, See-Kiong Ng, Patrick Jaillet, and Bryan Kian~Hsiang Low.
\newblock Use your {INSTINCT}: {INST}ruction optimization for {LLM}s using neural bandits coupled with transformers.
\newblock In {\em Forty-first International Conference on Machine Learning}, 2024.

\bibitem{2024_Chen2}
Lichang Chen, Jiuhai Chen, Tom Goldstein, Heng Huang, and Tianyi Zhou.
\newblock {InstructZero}: Efficient instruction optimization for black-box large language models.
\newblock In {\em Forty-first International Conference on Machine Learning}, 2024.

\bibitem{2023_Zhou}
Yongchao Zhou, Andrei~Ioan Muresanu, Ziwen Han, Keiran Paster, Silviu Pitis, Harris Chan, and Jimmy Ba.
\newblock Large language models are human-level prompt engineers.
\newblock In {\em The Eleventh International Conference on Learning Representations}, 2023.

\bibitem{2023_Pryzant}
Reid Pryzant, Dan Iter, Jerry Li, Yin~Tat Lee, Chenguang Zhu, and Michael Zeng.
\newblock Automatic prompt optimization with ''gradient descent'' and beam search.
\newblock In {\em The 2023 Conference on Empirical Methods in Natural Language Processing}, 2023.

\bibitem{2024_Sabbatella}
Antonio Sabbatella, Andrea Ponti, Ilaria Giordani, Antonio Candelieri, and Francesco Archetti.
\newblock Prompt optimization in large language models.
\newblock {\em Mathematics}, 12(6):929, 2024.

\bibitem{2023_Chen2}
Yuyan Chen, Zhihao Wen, Ge~Fan, Zhengyu Chen, Wei Wu, Dayiheng Liu, Zhixu Li, Bang Liu, and Yanghua Xiao.
\newblock {MAPO}: Boosting large language model performance with model-adaptive prompt optimization.
\newblock In {\em The 2023 Conference on Empirical Methods in Natural Language Processing}, 2023.

\bibitem{2024_Wang3}
Xinyuan Wang, Chenxi Li, Zhen Wang, Fan Bai, Haotian Luo, Jiayou Zhang, Nebojsa Jojic, Eric Xing, and Zhiting Hu.
\newblock {PromptAgent}: Strategic planning with language models enables expert-level prompt optimization.
\newblock In {\em The Twelfth International Conference on Learning Representations}, 2024.

\bibitem{2024_Manas}
Oscar Ma{\~n}as, Pietro Astolfi, Melissa Hall, Candace Ross, Jack Urbanek, Adina Williams, Aishwarya Agrawal, Adriana Romero-Soriano, and Michal Drozdzal.
\newblock Improving text-to-image consistency via automatic prompt optimization.
\newblock {\em arXiv preprint arXiv:2403.17804}, 2024.

\bibitem{2023_Do}
Xuan~Long Do, Yiran Zhao, Hannah Brown, Yuxi Xie, James~Xu Zhao, Nancy~F Chen, Kenji Kawaguchi, Michael Shieh, and Junxian He.
\newblock Prompt optimization via adversarial in-context learning.
\newblock {\em arXiv preprint arXiv:2312.02614}, 2023.

\bibitem{2024_Sordoni}
Alessandro Sordoni, Eric Yuan, Marc-Alexandre C{\^o}t{\'e}, Matheus Pereira, Adam Trischler, Ziang Xiao, Arian Hosseini, Friederike Niedtner, and Nicolas Le~Roux.
\newblock Joint prompt optimization of stacked {LLMs} using variational inference.
\newblock {\em Advances in Neural Information Processing Systems}, 36, 2024.

\bibitem{2023_Sabbatella}
Antonio Sabbatella, Andrea Ponti, Antonio Candelieri, Ilaria Giordani, and Francesco Archetti.
\newblock A {Bayesian} approach for prompt optimization in pre-trained language models.
\newblock {\em arXiv preprint arXiv:2312.00471}, 2023.

\bibitem{2024_Wen3}
Yuxin Wen, Neel Jain, John Kirchenbauer, Micah Goldblum, Jonas Geiping, and Tom Goldstein.
\newblock Hard prompts made easy: Gradient-based discrete optimization for prompt tuning and discovery.
\newblock {\em Advances in Neural Information Processing Systems}, 36, 2024.

\bibitem{2023_Ye}
Qinyuan Ye, Maxamed Axmed, Reid Pryzant, and Fereshte Khani.
\newblock Prompt engineering a prompt engineer.
\newblock {\em arXiv preprint arXiv:2311.05661}, 2023.

\bibitem{2024_Wu3}
Shirley Wu, Shiyu Zhao, Qian Huang, Kexin Huang, Michihiro Yasunaga, Kaidi Cao, Vassilis~N Ioannidis, Karthik Subbian, Jure Leskovec, and James Zou.
\newblock {AvaTaR}: Optimizing {LLM} agents for tool-assisted knowledge retrieval.
\newblock {\em arXiv preprint arXiv:2406.11200}, 2024.

\bibitem{2024_Qu}
Changle Qu, Sunhao Dai, Xiaochi Wei, Hengyi Cai, Shuaiqiang Wang, Dawei Yin, Jun Xu, and Ji-Rong Wen.
\newblock Tool learning with large language models: A survey.
\newblock {\em arXiv preprint arXiv:2405.17935}, 2024.

\bibitem{2024_Schick}
Timo Schick, Jane Dwivedi-Yu, Roberto Dess{\`\i}, Roberta Raileanu, Maria Lomeli, Eric Hambro, Luke Zettlemoyer, Nicola Cancedda, and Thomas Scialom.
\newblock Toolformer: Language models can teach themselves to use tools.
\newblock {\em Advances in Neural Information Processing Systems}, 36, 2024.

\bibitem{2024_Qin}
Yujia Qin, Shengding Hu, Yankai Lin, Weize Chen, Ning Ding, Ganqu Cui, Zheni Zeng, Yufei Huang, Chaojun Xiao, Chi Han, et~al.
\newblock Tool learning with foundation models.
\newblock {\em arXiv preprint arXiv.2304.08354}, 10, 2023.

\bibitem{2024_Lumos}
Da~Yin, Faeze Brahman, Abhilasha Ravichander, Khyathi Chandu, Kai-Wei Chang, Yejin Choi, and Bill~Yuchen Lin.
\newblock Agent {Lumos}: Unified and modular training for open-source language agents.
\newblock In Lun-Wei Ku, Andre Martins, and Vivek Srikumar, editors, {\em Proceedings of the 62nd Annual Meeting of the Association for Computational Linguistics (Volume 1: Long Papers)}, pages 12380--12403, Bangkok, Thailand, August 2024. Association for Computational Linguistics.

\bibitem{2024_Zhou}
Wangchunshu Zhou, Yixin Ou, Shengwei Ding, Long Li, Jialong Wu, Tiannan Wang, Jiamin Chen, Shuai Wang, Xiaohua Xu, Ningyu Zhang, et~al.
\newblock Symbolic learning enables self-evolving agents.
\newblock {\em arXiv preprint arXiv:2406.18532}, 2024.

\bibitem{2024_Hu}
Shengran Hu, Cong Lu, and Jeff Clune.
\newblock Automated design of agentic systems.
\newblock {\em arXiv preprint arXiv:2408.08435}, 2024.

\bibitem{2022_Khattab}
Omar Khattab, Keshav Santhanam, Xiang~Lisa Li, David Hall, Percy Liang, Christopher Potts, and Matei Zaharia.
\newblock Demonstrate-search-predict: Composing retrieval and language models for knowledge-intensive {NLP}.
\newblock {\em arXiv preprint arXiv:2212.14024}, 2022.

\bibitem{2023_Singhvi}
Arnav Singhvi, Manish Shetty, Shangyin Tan, Christopher Potts, Koushik Sen, Matei Zaharia, and Omar Khattab.
\newblock {DSPy} assertions: Computational constraints for self-refining language model pipelines.
\newblock {\em arXiv preprint arXiv:2312.13382}, 2023.

\bibitem{2024_Khattab}
Omar Khattab, Arnav Singhvi, Paridhi Maheshwari, Zhiyuan Zhang, Keshav Santhanam, Saiful Haq, Ashutosh Sharma, Thomas~T Joshi, Hanna Moazam, Heather Miller, et~al.
\newblock {DSPy: Compiling Declarative Language Model Calls into State-of-the-Art Pipelines}.
\newblock In {\em The Twelfth International Conference on Learning Representations}, 2024.

\bibitem{2024_OpenR}
Jun Wang, Meng Fang, Ziyu Wan, Muning Wen, Jiachen Zhu, Anjie Liu, Ziqin Gong, Yan Song, Lei Chen, Lionel~M Ni, et~al.
\newblock {OpenR}: An open source framework for advanced reasoning with large language models.
\newblock {\em arXiv preprint arXiv:2410.09671}, 2024.

\bibitem{2024_Huang}
Qian Huang, Jian Vora, Percy Liang, and Jure Leskovec.
\newblock {MLAgentBench}: Evaluating language agents on machine learning experimentation.
\newblock In {\em Forty-first International Conference on Machine Learning}, 2024.

\bibitem{2024_Guo_ds}
Siyuan Guo, Cheng Deng, Ying Wen, Hechang Chen, Yi~Chang, and Jun Wang.
\newblock {DS-Agent}: Automated data science by empowering large language models with case-based reasoning.
\newblock {\em arXiv preprint arXiv:2402.17453}, 2024.

\bibitem{2024_Grosnit}
Antoine Grosnit, Alexandre Maraval, James Doran, Giuseppe Paolo, Albert Thomas, Refinath Shahul Hameed~Nabeezath Beevi, Jonas Gonzalez, Khyati Khandelwal, Ignacio Iacobacci, Abdelhakim Benechehab, et~al.
\newblock Large language models orchestrating structured reasoning achieve {Kaggle} grandmaster level.
\newblock {\em arXiv preprint arXiv:2411.03562}, 2024.

\bibitem{2024_Hu3}
Xueyu Hu, Ziyu Zhao, Shuang Wei, Ziwei Chai, Qianli Ma, Guoyin Wang, Xuwu Wang, Jing Su, Jingjing Xu, Ming Zhu, et~al.
\newblock {InfiAgent-DABench}: Evaluating agents on data analysis tasks.
\newblock In {\em Forty-first International Conference on Machine Learning}, 2024.

\bibitem{2024_Li}
Jinyang Li, Nan Huo, Yan Gao, Jiayi Shi, Yingxiu Zhao, Ge~Qu, Yurong Wu, Chenhao Ma, Jian-Guang Lou, and Reynold Cheng.
\newblock {Tapilot-Crossing}: Benchmarking and evolving llms towards interactive data analysis agents.
\newblock {\em arXiv preprint arXiv:2403.05307}, 2024.

\bibitem{2024_Liu}
Xiao Liu, Zirui Wu, Xueqing Wu, Pan Lu, Kai-Wei Chang, and Yansong Feng.
\newblock Are {LLMs} capable of data-based statistical and causal reasoning? benchmarking advanced quantitative reasoning with data.
\newblock {\em arXiv preprint arXiv:2402.17644}, 2024.

\bibitem{2023_Jin}
Zhijing Jin, Yuen Chen, Felix Leeb, Luigi Gresele, Ojasv Kamal, Zhiheng LYU, Kevin Blin, Fernando~Gonzalez Adauto, Max Kleiman-Weiner, Mrinmaya Sachan, and Bernhard Sch{\"o}lkopf.
\newblock {CL}adder: A benchmark to assess causal reasoning capabilities of language models.
\newblock In {\em Thirty-seventh Conference on Neural Information Processing Systems}, 2023.

\bibitem{2024_Majumder}
Bodhisattwa~Prasad Majumder, Harshit Surana, Dhruv Agarwal, Bhavana~Dalvi Mishra, Abhijeetsingh Meena, Aryan Prakhar, Tirth Vora, Tushar Khot, Ashish Sabharwal, and Peter Clark.
\newblock {DiscoveryBench}: Towards data-driven discovery with large language models.
\newblock {\em arXiv preprint arXiv:2407.01725}, 2024.

\bibitem{2024_Mirza}
Adrian Mirza, Nawaf Alampara, Sreekanth Kunchapu, Benedict Emoekabu, Aswanth Krishnan, Mara Wilhelmi, Macjonathan Okereke, Juliane Eberhardt, Amir~Mohammad Elahi, Maximilian Greiner, et~al.
\newblock Are large language models superhuman chemists?
\newblock {\em arXiv preprint arXiv:2404.01475}, 2024.

\bibitem{2024_Blade}
Ken Gu, Ruoxi Shang, Ruien Jiang, Keying Kuang, Richard-John Lin, Donghe Lyu, Yue Mao, Youran Pan, Teng Wu, Jiaqian Yu, et~al.
\newblock {BLADE}: Benchmarking language model agents for data-driven science.
\newblock In {\em Empirical Methods in Natural Language Processing}, 2024.

\bibitem{2024_Ma}
Yubo Ma, Zhibin Gou, Junheng Hao, Ruochen Xu, Shuohang Wang, Liangming Pan, Yujiu Yang, Yixin Cao, and Aixin Sun.
\newblock {SciAgent}: Tool-augmented language models for scientific reasoning.
\newblock In {\em Proceedings of the 2024 Conference on Empirical Methods in Natural Language Processing}, 2024.

\bibitem{2024_Jansen}
Peter Jansen, Marc-Alexandre C{\^o}t{\'e}, Tushar Khot, Erin Bransom, Bhavana~Dalvi Mishra, Bodhisattwa~Prasad Majumder, Oyvind Tafjord, and Peter Clark.
\newblock {DISCOVERYWORLD: A Virtual Environment for Developing and Evaluating Automated Scientific Discovery Agents}.
\newblock In {\em Advances in Neural Information Processing Systems}, 2024.

\bibitem{2022_Wang}
Ruoyao Wang, Peter Jansen, Marc-Alexandre C{\^o}t{\'e}, and Prithviraj Ammanabrolu.
\newblock {ScienceWorld}: Is your agent smarter than a 5th grader?
\newblock In {\em Proceedings of the 2022 Conference on Empirical Methods in Natural Language Processing}, pages 11279--11298, 2022.

\bibitem{ramos2024review}
Mayk~Caldas Ramos, Christopher Collison, and Andrew~D White.
\newblock A review of large language models and autonomous agents in chemistry.
\newblock {\em Chemical Science}, 2024.

\bibitem{1965_Aastrom}
Karl~Johan {\AA}str{\"o}m.
\newblock {Optimal control of Markov processes with incomplete state information I}.
\newblock {\em Journal of Mathematical Analysis and Applications}, 10:174--205, 1965.

\bibitem{1956_Kleene}
SC~Kleene.
\newblock Representation of events in nerve nets and finite automata.
\newblock {\em Automata Studies: Annals of Mathematics Studies. Number 34}, 34:3, 1956.

\bibitem{2023_Meister}
Clara Meister, Tiago Pimentel, Gian Wiher, and Ryan Cotterell.
\newblock Locally typical sampling.
\newblock {\em Transactions of the Association for Computational Linguistics}, 11:102--121, 2023.

\bibitem{2024_Wu}
Yue Wu, Yewen Fan, Paul~Pu Liang, Amos Azaria, Yuanzhi Li, and Tom~M Mitchell.
\newblock Read and reap the rewards: Learning to play {Atari} with the help of instruction manuals.
\newblock {\em Advances in Neural Information Processing Systems}, 36, 2024.

\bibitem{1964_Widrow}
B~Widrow and FW~Smith.
\newblock {\em Computer and Information Sciences}, chapter Pattern recognising control systems.
\newblock Clever Hume Press, 1964.

\bibitem{1969_Chambers}
Roger~A Chambers and Donald Michie.
\newblock Man-machine co-operation on a learning task.
\newblock {\em Computer Graphics: Techniques and Applications}, pages 179--186, 1969.

\bibitem{1988_Pomerleau}
Dean~A Pomerleau.
\newblock {ALVINN}: An autonomous land vehicle in a neural network.
\newblock {\em Advances in neural information processing systems}, 1, 1988.

\bibitem{1990_Michie}
D.~Michie, M.~Bain, and J.~Hayes-Michie.
\newblock {\em Cognitive models from subcognitive skills}, chapter Chapter 5, pages 71--99.
\newblock The Institution of Engineering and Technology, 1990.

\bibitem{1995_Bain}
Michael Bain and Claude Sammut.
\newblock A framework for behavioural cloning.
\newblock {\em Machine Intelligence}, 15:103--129, 1995.

\bibitem{2023_Shunyu}
Baian Chen, Chang Shu, Ehsan Shareghi, Nigel Collier, Karthik Narasimhan, and Shunyu Yao.
\newblock {FireAct}: Toward language agent fine-tuning.
\newblock {\em arXiv preprint arXiv:2310.05915}, 2023.

\bibitem{2023_Zeng}
Aohan Zeng, Mingdao Liu, Rui Lu, Bowen Wang, Xiao Liu, Yuxiao Dong, and Jie Tang.
\newblock {AgentTuning}: Enabling generalized agent abilities for {LLMs}.
\newblock {\em arXiv preprint arXiv:2310.12823}, 2023.

\bibitem{2023_Yuan2}
Zheng Yuan, Hongyi Yuan, Chengpeng Li, Guanting Dong, Keming Lu, Chuanqi Tan, Chang Zhou, and Jingren Zhou.
\newblock Scaling relationship on learning mathematical reasoning with large language models.
\newblock {\em arXiv preprint arXiv:2308.01825}, 2023.

\bibitem{brown2024large}
Bradley Brown, Jordan Juravsky, Ryan Ehrlich, Ronald Clark, Quoc~V Le, Christopher R{\'e}, and Azalia Mirhoseini.
\newblock Large language monkeys: Scaling inference compute with repeated sampling.
\newblock {\em arXiv preprint arXiv:2407.21787}, 2024.

\bibitem{wang2022self}
Xuezhi Wang, Jason Wei, Dale Schuurmans, Quoc~V Le, Ed~H. Chi, Sharan Narang, Aakanksha Chowdhery, and Denny Zhou.
\newblock Self-consistency improves chain of thought reasoning in language models.
\newblock In {\em The Eleventh International Conference on Learning Representations}, 2023.

\bibitem{li2022competition}
Yujia Li, David Choi, Junyoung Chung, Nate Kushman, Julian Schrittwieser, R{\'e}mi Leblond, Tom Eccles, James Keeling, Felix Gimeno, Agustin Dal~Lago, et~al.
\newblock Competition-level code generation with alphacode.
\newblock {\em Science}, 378(6624):1092--1097, 2022.

\bibitem{tanitvy-code}
Quickwit Inc.
\newblock {tantivy}, October 2024.

\bibitem{Bertero2017Methods}
Alessandro Bertero, Stephanie Brown, and Ludovic Vallier.
\newblock {\em Methods of Cloning}, page 19–39.
\newblock Elsevier, 2017.

\bibitem{Sharma2014Advances}
Kamal Sharma, Ajay~Kumar Mishra, Vikram Mehraj, and Ganesh~Selvaraj Duraisamy.
\newblock Advances and applications of molecular cloning in clinical microbiology.
\newblock {\em Biotechnology and Genetic Engineering Reviews}, 30(1):65–78, 2014.

\bibitem{steinegger2017mmseqs2}
Martin Steinegger and Johannes S{\"o}ding.
\newblock {MMseqs2} enables sensitive protein sequence searching for the analysis of massive data sets.
\newblock {\em Nature Biotechnology}, 35(11):1026--1028, 2017.

\bibitem{sheldon2018role}
Roger~A Sheldon and John~M Woodley.
\newblock Role of biocatalysis in sustainable chemistry.
\newblock {\em Chemical reviews}, 118(2):801--838, 2018.

\bibitem{Goldenzweig2018-rm}
Adi Goldenzweig and Sarel~J Fleishman.
\newblock Principles of protein stability and their application in computational design.
\newblock {\em Annu. Rev. Biochem.}, 87:105--129, June 2018.

\bibitem{thermompnn}
Henry Dieckhaus, Michael Brocidiacono, Nicholas~Z. Randolph, and Brian Kuhlman.
\newblock Transfer learning to leverage larger datasets for improved prediction of protein stability changes.
\newblock {\em Proceedings of the National Academy of Sciences}, 121(6):e2314853121, 2024.

\bibitem{BROOM2020717}
Aron Broom, Kyle Trainor, Zachary Jacobi, and Elizabeth~M. Meiering.
\newblock Computational modeling of protein stability: Quantitative analysis reveals solutions to pervasive problems.
\newblock {\em Structure}, 28(6):717--726.e3, 2020.

\bibitem{MAESTRO}
Josef Laimer, Heidi Hofer, Marko Fritz, Stefan Wegenkittl, and Peter Lackner.
\newblock {MAESTRO--multi} agent stability prediction upon point mutations.
\newblock {\em BMC Bioinformatics}, 16(1):116, April 2015.

\bibitem{sapscore}
Timothy~M Lauer, Neeraj~J Agrawal, Naresh Chennamsetty, Kamal Egodage, Bernhard Helk, and Bernhardt~L Trout.
\newblock Developability index: a rapid in silico tool for the screening of antibody aggregation propensity.
\newblock {\em J. Pharm. Sci.}, 101(1):102--115, January 2012.

\bibitem{310ai}
310.ai.
\newblock 310 copilot.
\newblock 2024.

\bibitem{ghafarollahi_protagents_2024}
Alireza Ghafarollahi and Markus~J Buehler.
\newblock {ProtAgents}: protein discovery via large language model multi-agent collaborations combining physics and machine learning.
\newblock {\em Digital Discovery}, 2024.

\bibitem{ingraham2023illuminating}
John~B. Ingraham, Maxim Baranov, Zak Costello, et~al.
\newblock Illuminating protein space with a programmable generative model.
\newblock {\em Nature}, 623:1070--1078, 2023.

\bibitem{Wu2022.07.21.500999}
Ruidong Wu, Fan Ding, Rui Wang, Rui Shen, Xiwen Zhang, Shitong Luo, Chenpeng Su, Zuofan Wu, Qi~Xie, Bonnie Berger, Jianzhu Ma, and Jian Peng.
\newblock High-resolution de novo structure prediction from primary sequence.
\newblock {\em bioRxiv}, 2022.

\bibitem{chen2024llmshighlyconstrainedbiophysicalsequence}
Angelica Chen, Samuel~D Stanton, Robert~G Alberstein, Andrew~M Watkins, Richard Bonneau, Vladimir Gligorijevi, Kyunghyun Cho, and Nathan~C Frey.
\newblock {LLMs} are highly-constrained biophysical sequence optimizers.
\newblock {\em arXiv preprint arXiv:2410.22296}, 2024.

\bibitem{Tsuboyama2023}
Kotaro Tsuboyama, Justas Dauparas, Jonathan Chen, Elodie Laine, Yasser Mohseni~Behbahani, Jonathan~J. Weinstein, Niall~M. Mangan, Sergey Ovchinnikov, and Gabriel~J. Rocklin.
\newblock Mega-scale experimental analysis of protein folding stability in biology and design.
\newblock {\em Nature}, 620(7973):434--444, Aug 2023.

\bibitem{Frenz2020}
Brandon Frenz, Steven~M Lewis, Indigo King, Frank DiMaio, Hahnbeom Park, and Yifan Song.
\newblock Prediction of protein mutational free energy: Benchmark and sampling improvements increase classification accuracy.
\newblock {\em Front. Bioeng. Biotechnol.}, 8:558247, October 2020.

\bibitem{anthropic2024claude3}
Anthropic.
\newblock Introducing the next generation of {Claude}, 2024.
\newblock https:\slash\slash www.anthropic.com\slash news\slash claude-3-family.

\bibitem{mirzadeh2024}
Iman Mirzadeh, Keivan Alizadeh, Hooman Shahrokhi, Oncel Tuzel, Samy Bengio, and Mehrdad Farajtabar.
\newblock {GSM}-symbolic: Understanding the limitations of mathematical reasoning in large language models.
\newblock {\em arXiv preprint arXiv:2410.05229}, 2024.

\bibitem{grattafiori2024llama3herdmodels}
Aaron Grattafiori, Abhimanyu Dubey, Abhinav Jauhri, Abhinav Pandey, Abhishek Kadian, Ahmad Al-Dahle, and Aiesha~Letman et. al.
\newblock The {Llama} 3 herd of models, 2024.

\bibitem{joint_aisi}
UK~AI Safety~Institute US~AI Safety~Institute.
\newblock Pre-deployment evaluation of {Anthropic’s} upgraded {Claude 3.5 Sonnet}.
\newblock {\em Technical Report}, 2024.

\bibitem{srivastava2023beyond}
BIG bench authors.
\newblock Beyond the imitation game: Quantifying and extrapolating the capabilities of language models.
\newblock {\em Transactions on Machine Learning Research}, 2023.

\end{thebibliography}

\clearpage
\appendix
\pagebreak
\renewcommand{\thefigure}{A\arabic{figure}}
\setcounter{figure}{0}

\renewcommand{\thetable}{A\arabic{table}}
\setcounter{table}{0}
\section{Environment Details}
\begin{longtable}{|p{0.25\textwidth}|p{0.75\textwidth}|}
\caption{Environments implemented within the Aviary framework.} \\

\hline
\rowcolor{gray!30} \multicolumn{2}{|c|}{\textbf{PaperQA}} \\
\hline 
Task & LitQA2 questions \\
\hline
Example Task Element &

Q: Which base editor has been shown to be the most efficient for inducing the mutation K352E in CD45 in human T-cells?
\newline \newline Options:
\newline A) ABE8e-NG
\newline B) ABE8e–SpRY
\newline C) SPACE-NG
\newline D) Insufficient information to answer this question
\newline E) ABE8e-SpG
\newline \newline  Answer: \\
\hline
Tools &

\begin{itemize}[leftmargin=*]
    \item \texttt{paper\_search(query:\;str, min\_year:\;int\:|\:None, max\_year:\;int\:|\:None)} - Full-text semantic search through a local search index.
    \item \texttt{gather\_evidence(question:\;str)} - Perform LLM reranking and contextual summarization given a question on \texttt{paper\_search} results.
    \item \texttt{gen\_answer()} - Attempt to answer given the top ranked contextual summaries.
    \item \texttt{complete(has\_successful\_answer:\;bool)} - Terminate using the last proposed answer, with the argument declaring if the answer addressed all parts of the question.
\end{itemize}

\\
\hline
Reward $R$ & 

\begin{equation*}
    \begin{cases} 
    1 & \text{correct answer}, \\
    -1 & \text{incorrect answer}, \\
    0.1 & \text{unsure answer}.
    \end{cases}
\end{equation*}

\\
\hline
Transition Function $\mathcal{T}$ & Nondeterministic, due to the stochastic nature of LLM prompting within both the \texttt{gather\_evidence} and \texttt{gen\_answer} tools. \\
\hline
Code & \href{https://pypi.org/project/paper-qa/5.6.1/}{\texttt{paper-qa==5.6.1}} \\
\hline
\hline
\rowcolor{gray!30} \multicolumn{2}{|c|}{\textbf{\textsc{hotpot}QA}} \\
\hline
Task & \textsc{hotpot}QA questions \\
\hline
Example Task Element & A robe takes 2 bolts of blue fiber and half that much white fiber.  How many bolts in total does it take? \\
\hline
Tools &

\begin{itemize}[leftmargin=*]
    \item \texttt{search(entity:\;str)} - Search Wikipedia for an entity, keeping either matching sentences or most similar pages.
    \item \texttt{lookup(keyword:\;str)} - Keyword lookup within all search results, emulating the ``find'' functionality of a web browser.
    \item \texttt{submit\_answer(answer:\;str)} - Check if an answer is correct.
\end{itemize}

\\
\hline
Reward $R$ &

\begin{equation*}
    \begin{cases} 
    1 & \text{correct answer}, \\
    0 & \text{otherwise}.
    \end{cases}
\end{equation*}

\\
\hline
Transition Function $\mathcal{T}$ & Deterministic. \\
\hline
Code & \href{https://pypi.org/project/aviary.gsm8k/0.11.0/}{\texttt{aviary.gsm8k==0.11.0}} \\
\hline
\hline
\rowcolor{gray!30} \multicolumn{2}{|c|}{\textbf{GSM8k}} \\
\hline
Task & GSM8k questions \\
\hline
Example Task Element & Weng earns \$12 an hour for babysitting. Yesterday, she just did 50 minutes of babysitting. How much did she earn? \\
\hline
Tools &

\begin{itemize}[leftmargin=*]
    \item \texttt{calculator(expr:\;str)} - Return the result of a numerical expression.
    \item \texttt{submit\_answer(answer:\;str)} - Check if an answer is correct.
\end{itemize}

\\
\hline
Reward $R$ & 

\begin{equation*}
    \begin{cases} 
    1 & \text{correct answer}, \\
    -1 & \text{invalid tool call}, \\
    0 & \text{otherwise}.
    \end{cases}
\end{equation*}

\\
\hline
Transition Function $\mathcal{T}$ & Deterministic. \\
\hline
Code & \href{https://pypi.org/project/aviary.hotpotqa/0.11.0/}{\texttt{aviary.hotpotqa==0.11.0}} \\
\hline
\hline
\rowcolor{gray!30} \multicolumn{2}{|c|}{\textbf{Molecular Cloning}} \\
\hline
Task & SeqQA questions \\
\hline
Example Task Element & 
 Q: Which of the following RNA sequences contains an ORF that is most likely to have high translation efficiency in a human cell?
\newline \newline Options:
\newline A) [RNA sequence]
\newline B) [RNA sequence]
\newline C) Insufficient information to answer this question
\newline D) [RNA sequence]
\newline E) [RNA sequence] \\
\hline
Tools &

\begin{itemize}[leftmargin=*]
    \item \texttt{search(query:\;str)} - Search Plasmid and NCBI nucleotide databases.
    \item \texttt{annotate(sequence:\;Sequence)} - Annotate proteins, ORFs, restriction sites in a DNA sequence.
    \item \texttt{gibson(sequences:\;list[Sequence])} - Simulate gibson assembly.
    \item \texttt{goldengate(sequences:\;list[Sequence], enzyme:\;str)} - Simulate golden gate assembly/
    \item \texttt{simulate\_pcr(sequence:\;Sequence, forward\_primer:\;Sequence\:|\:None, forward\_primer\_name:\;str\:|\:None)} - Simulate polymerase chain reaction.
    \item \texttt{optimize\_translation(sequence:\:Sequence, cg\_content:\;int, codon\_table:\;int, min\_repeat\_length:\;int)} - Codon optimization.
    \item \texttt{separate(sequences:\;list[Sequence])} - Simulate gel electrophoresis.
    \item \texttt{enzyme\_cut(sequence:\;Sequence, enzyme:\;str)} - Simulate restriction digest.
    \item \texttt{search(query:\;str)} - Search Plasmid and NCBI nucleotide databases.
    \item \texttt{annotate(sequence:\;Sequence)} - Annotate proteins, ORFs, restriction sites in a DNA sequence.
    \item \texttt{gibson(sequences:\;list[Sequence])} - Simulate gibson assembly.
    \item \texttt{goldengate(sequences:\;list[Sequence], enzyme:\;str)} - Simulate golden gate assembly. 
    \item \texttt{simulate\_pcr(sequence:\;Sequence, forward\_primer:\;Sequence\:|\:None, forward\_primer\_name:\;str\:|\:None)} - Simulate polymerase chain reaction. Primers can be sequence (by ref) or name of enzyme or sequence value.
    \item \texttt{optimize\_translation(sequence:\;Sequence, cg\_content:\;int, codon\_table:\;int, min\_repeat\_length:\;int)} - Codon optimization.
    \item \texttt{separate(sequences:\;list[Sequence])} - Simulate gel electrophoresis.
    \item \texttt{enzyme\_cut(sequence:\;Sequence, enzyme:\;str)} - Simulate restriction digest.
    \item \texttt{find\_sequence\_overlap(sequence1:\;Sequence, sequence2:\;Sequence, reverse:\;bool)} - Find overlapping regions between two sequences.
    \item \texttt{find\_orfs(sequence:\;Sequence, min\_length:\;int, codon\_table:\;int, strand:\;int)} - Find open reading frames in a DNA sequence.
    \item \texttt{design\_primers(sequence:\;Sequence, target\_tm:\;float, forward\_overhang\_name:\;str, reverse\_overhang\_name:\;str)} - Design PCR primers for a sequence.
    \item \texttt{merge(sequences:\;list[Sequence])} - Combine multiple sequences, needed to do assembly simulation.
    \item \texttt{add(sequence1:\;Sequence, sequence2:\;Sequence)} - Add two sequences together.
    \item \texttt{slice\_sequence(sequence:\;Sequence, start:\;int, end:\;int, name:\;str)} - Extract a subsequence.
    \item \texttt{view\_translation(sequence:\;Sequence)} - View the amino acid translation of a DNA sequence.
    \item \texttt{view\_sequence\_stats(sequence:\;Sequence)} - View sequence statistics.
    \item \texttt{view\_restriction\_sites(sequence:\;Sequence)} - View restriction enzyme cut sites.
    \item \texttt{view\_sequence(sequence:\;Sequence)} - View the raw sequence.
    \item \texttt{submit\_answer(answer:\;str)}
\end{itemize}

\\
\hline

Reward $R$ & \begin{equation*}
    \begin{cases} 
    1 & \text{correct answer}, \\
    -1 & \text{incorrect answer}, \\
    0.1 & \text{unsure answer}.
    \end{cases}
\end{equation*}

\\
\hline
Transition Function $\mathcal{T}$ & Deterministic. \\
\hline
\hline
\rowcolor{gray!30} \multicolumn{2}{|c|}{\textbf{Protein Design}} \\
\hline
Task & Protein Stability \\
\hline
Example Task & Design at least 3 mutations and a maximum of 7 mutations to the protein sequence \{sequence\} that would improve its 
stability. The sequence of this protein is provided in the text file located at \{local\_txt\_file\}, and the structure of the protein can be found in the PDB file located at \{local\_pdb\_file\}. \\
\hline
Tools & \begin{itemize}[leftmargin=*]
    \item \texttt{get\_bond\_types\_between(residues:\;list[int], bond\_type:\;str)} - Describes all instances of the specified bond type among a given list of residues as outlined in the function description.
    \item \texttt{get\_secondary\_structure(pdb\_string:\;str)}
    - Describes secondary structure elements found in the protein structure by residue.
    \item \texttt{get\_sequence\_properties(mutations:\;list[str], return\_wt:\;bool)} - Describes properties like instability index, molar extinction coefficient, fraction of charged residues, iso-electric point.
    \item \texttt{get\_distance\_between\_residues(mutation:\;list[str])} - Get pairwise distances between list of residues.
    \item \texttt{get\_residue\_at\_position(residues:\;list[int])} - Returns the residue present at a specific position and describes whether it is acidic or basic or charged, polar or aliphatic or aromatic.
    \item \texttt{get\_hydrophobicity\_score(local\_pdb\_file:\;str)} - Calculates aggregation propensity by  residue using Rosetta.
    \item\texttt{get\_mutant\_protein\_sequence(mutations:\;list[str])} - Returns the sequence of the protein after the mutations are applied to the sequence.
\end{itemize} \\
\hline
Reward $R$ & \begin{equation*}
    \begin{cases} 
    1 & \text{RosettaddG < 0}, \\
    0 & \text{otherwise.}
    \end{cases}
\end{equation*} \\
\hline
Transition Function $\mathcal{T}$ & Deterministic. \\
\hline
\end{longtable}

\end{document}